\pdfoutput=1
\documentclass{article}
\usepackage{microtype}
\usepackage{graphicx}
\usepackage{booktabs,tabularx,multirow}
\usepackage[hyperfootnotes=false]{hyperref}
\usepackage{xcolor,url}
\usepackage{enumitem}
\usepackage[preprint]{icml2026}
\hypersetup{pdfsubject={Preprint}}
\usepackage{amsmath,amssymb,mathtools}
\usepackage{etoolbox}
\makeatletter
\let\paperOriginalSection\section
\let\paperOriginalParagraph\paragraph
\makeatother
\newlength{\paperTextFloatSep}
\newlength{\paperDblTextFloatSep}
\newlength{\paperFloatSep}
\newlength{\paperDblFloatSep}
\newlength{\paperInTextSep}
\newlength{\paperAboveDisplaySkip}
\newlength{\paperBelowDisplaySkip}
\newlength{\paperAboveDisplayShortSkip}
\newlength{\paperBelowDisplayShortSkip}
\newcommand{\compactMainLayout}{%
  \setlength{\paperTextFloatSep}{\textfloatsep}%
  \setlength{\paperDblTextFloatSep}{\dbltextfloatsep}%
  \setlength{\paperFloatSep}{\floatsep}%
  \setlength{\paperDblFloatSep}{\dblfloatsep}%
  \setlength{\paperInTextSep}{\intextsep}%
  \setlength{\paperAboveDisplaySkip}{\abovedisplayskip}%
  \setlength{\paperBelowDisplaySkip}{\belowdisplayskip}%
  \setlength{\paperAboveDisplayShortSkip}{\abovedisplayshortskip}%
  \setlength{\paperBelowDisplayShortSkip}{\belowdisplayshortskip}%
  \setlength{\textfloatsep}{8pt plus 2pt minus 1pt}%
  \setlength{\dbltextfloatsep}{8pt plus 2pt minus 1pt}%
  \setlength{\floatsep}{6pt plus 1pt minus 1pt}%
  \setlength{\dblfloatsep}{6pt plus 1pt minus 1pt}%
  \setlength{\intextsep}{7pt plus 1pt minus 1pt}%
  \setlength{\abovedisplayskip}{5pt plus 1pt minus 1pt}%
  \setlength{\belowdisplayskip}{5pt plus 1pt minus 1pt}%
  \setlength{\abovedisplayshortskip}{3pt plus 1pt}%
  \setlength{\belowdisplayshortskip}{3pt plus 1pt}%
  \captionsetup{skip=4pt}%
  \patchcmd{\section}{-0.12in}{-5pt}{}{\PackageError{paper}{Section spacing patch failed}{}}%
  \patchcmd{\paragraph}{1.5ex plus 0.5ex minus .2ex}{2pt plus 1pt minus 1pt}{}{\PackageError{paper}{Paragraph spacing patch failed}{}}%
  \raggedbottom
}
\newcommand{\restorePaperLayout}{%
  \let\section\paperOriginalSection
  \let\paragraph\paperOriginalParagraph
  \setlength{\textfloatsep}{\paperTextFloatSep}%
  \setlength{\dbltextfloatsep}{\paperDblTextFloatSep}%
  \setlength{\floatsep}{\paperFloatSep}%
  \setlength{\dblfloatsep}{\paperDblFloatSep}%
  \setlength{\intextsep}{\paperInTextSep}%
  \setlength{\abovedisplayskip}{\paperAboveDisplaySkip}%
  \setlength{\belowdisplayskip}{\paperBelowDisplaySkip}%
  \setlength{\abovedisplayshortskip}{\paperAboveDisplayShortSkip}%
  \setlength{\belowdisplayshortskip}{\paperBelowDisplayShortSkip}%
  \captionsetup{skip=0.1in}%
  \flushbottom
}

\newcommand{\R}{\mathbb{R}}

\icmltitlerunning{Transformers Stop Thinking Too Early, and a Tiny LoRA Fixes It}
\begin{document}
\compactMainLayout
\twocolumn[
\icmltitle{Transformers Stop Thinking Too Early, and a Tiny LoRA Fixes It}
\begin{icmlauthorlist}
\icmlauthor{Zehao Jin}{gatech}
\icmlauthor{Ruixuan Deng}{gatech}
\icmlauthor{Junran Wang}{gatech}
\end{icmlauthorlist}
\icmlaffiliation{gatech}{Georgia Institute of Technology}
\icmlcorrespondingauthor{Zehao Jin}{zehao@gatech.edu}
\icmlkeywords{transformers, low-rank adaptation, multi-hop reasoning, recurrent depth}
\vskip 0.15in
\begin{center}
\expandafter\def\csname @captype\endcsname{figure}
\captionsetup{hypcap=false}
{\includegraphics[width=\textwidth]{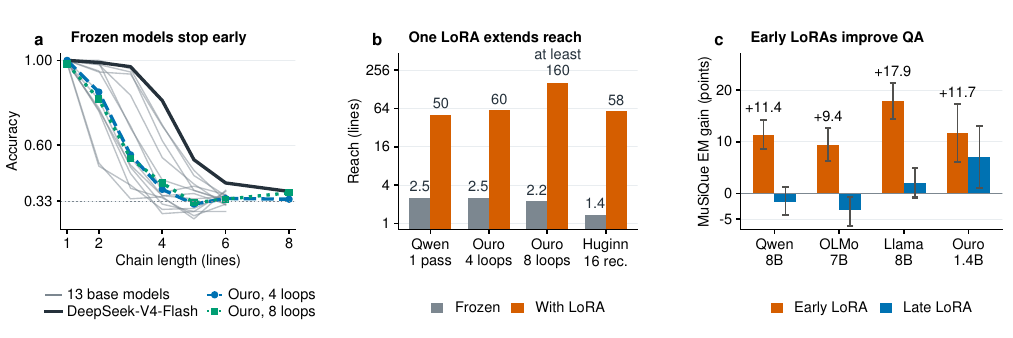}}
\caption{\textbf{Short reference chains become long computations after a tiny edit.} (a) Frozen three-chain choice accuracy. (b) Frozen and LoRA-enabled reach at 80\% accuracy, with the same metric within each pair: two-chain exact accuracy for Qwen; two-chain choice accuracy for Ouro and Huginn. (c) MuSiQue exact-match gains at the earliest and latest tested layers, with paired 95\% intervals. Both Ouro LoRAs act in every loop; a late LoRA still precedes the next loop's middle layers.}
\label{fig:teaser}
\end{center}
\vspace{0.12in}
]
\printAffiliationsAndNotice{}
\clubpenalty=10000
\widowpenalty=10000
\displaywidowpenalty=10000
\begin{abstract}
Pretrained transformers use little of their depth to follow references in context. Thirteen base models reliably follow only 1.4--3.6 lines, and extra pretrained loops add little. A task-trained rank-8 LoRA at one early layer extends this computation with all model weights frozen. Qwen3-8B improves from 15.5\% to 99\% exact accuracy on 24-line chains; a longer-trained LoRA reaches 50 lines. Ouro-1.4B reaches 60 lines after four loops and at least 160 after eight. The LoRA starts a relay: program lines pass on their chain identity through a short range of middle layers. Frozen heads read progressively further up the chain, and removing parent-line attention stops the relay. A frozen-model measurement locates the last useful intervention layer within tolerance in three of four held-out models. Task-specific LoRAs also improve MuSiQue. Default answers therefore understate the computation accessible through a tiny edit. Code and an interactive demo are available at \url{https://lunamos.github.io/stop-thinking-too-early/}.
\end{abstract}

\section{Introduction}
A transformer can read \texttt{K = apple}, \texttt{B = K}, \texttt{D = B}, and answer \texttt{print(D)} without retrieving any outside knowledge. Yet adding a few more assignments breaks this ability. Thirteen pretrained base models reliably follow only 1.4--3.6 lines (median 2.2). Reach grows slowly with size within families, yet even DeepSeek-V4-Flash (292B MoE) remains near chance on longer chains. Extra loops add little in pretrained looped models (Fig.~\ref{fig:teaser}a). These failures are surprising because transformers can implement much longer reference-following computations in principle \citep{sanford2024logdepth,saunshi2025latent}.

The failure is also easy to change. We train a rank-8 LoRA at one layer while freezing every model weight. In Qwen3-8B, this trains 65,537 added parameters, under 0.01\% of the model, and raises exact accuracy on 24-line chains from 15.5\% to 99\%. A LoRA trained on longer programs reaches 50 lines in one pass. In Ouro-1.4B, a LoRA makes additional loops useful: the model follows 60 lines after four loops and at least 160 after eight. LoRA on projection weights and FLAS-style flows at the same layer also work. Why can such a small change do so much? The key question is what the frozen layers do differently.

We answer that question by tracing the program's intermediate representations. By default, the program passes on which chain each line belongs to for only two or three lines. The query resolves another one or two pointers, and a late layer copies the selected root value. With the LoRA, the program's own lines carry this computation much further. Each line collects names from earlier lines and attends further up its chain. This \emph{relay} advances in a short range of middle layers, repeated in each loop. A LoRA applied only in Ouro's first loop already lets the unmodified later loops follow about 25 lines.

The mechanism suggests where adaptation can help: the intervention must act while useful middle-layer computation still follows it. Moving the same intervention past a model-specific limit removes most of its benefit. A measurement on the frozen model predicts this limit with modest precision on held-out models. On MuSiQue \citep{trivedi2022musique}, early LoRAs improve exact match, and projection LoRA restricted to early layers preserves most of the gain from adapting every layer.

Our contributions connect the failure, fix, and computation:
\begin{enumerate}[leftmargin=*,itemsep=0pt,topsep=2pt,parsep=0pt]
\item \textbf{A short default computation} across thirteen standard models and two looped families (\S\ref{sec:default}).
\item \textbf{A tiny edit with large gains:} 50-line reach in one pass and at least 160 with loops (\S\ref{sec:switch}).
\item \textbf{A causal account of the relay} and a prospective placement test, followed by task-specific adaptation on MuSiQue (\S\S\ref{sec:mechanism}--\ref{sec:downstream}).
\end{enumerate}

\begin{figure*}[t]
\centering
\includegraphics[width=\textwidth]{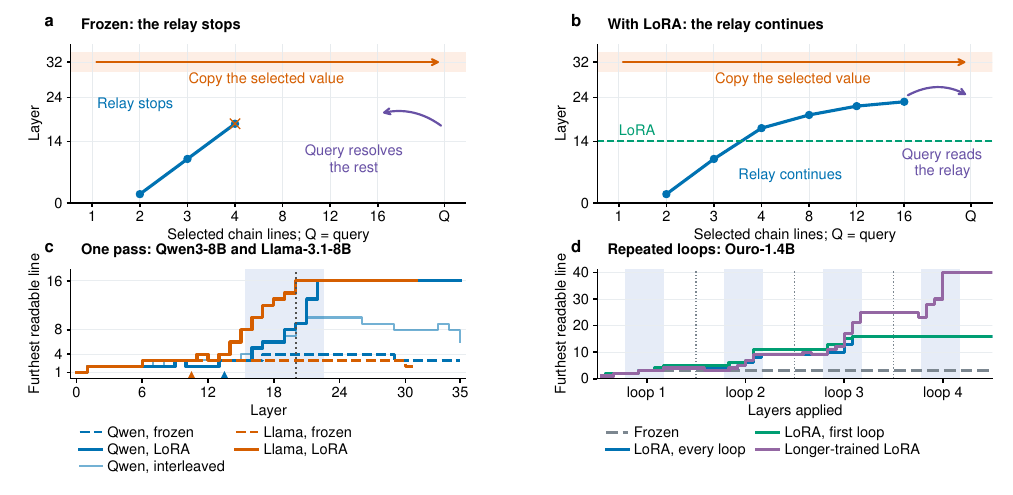}
\caption{\textbf{The LoRA extends a relay that runs in the same middle layers.} (a,b) Schematic of Qwen3-8B's default and LoRA-enabled computation: program lines carry chain identity; the query follows remaining pointers by default or reads the relay with the LoRA; late layers copy the value. Arrows summarize separate measurements, not a single trajectory. (c,d) The longest prefix of lines whose chain identity is decoded at 75\% accuracy: standard models across layers, and Ouro-1.4B across four loops. Programs have sixteen lines, except Ouro's longer-trained LoRA trace (forty). Triangles mark LoRA inputs; shading marks relay layers; the dotted line in (c) marks Qwen's placement limit.}
\label{fig:default_main}
\end{figure*}

\section{Task and intervention}\label{sec:setup}
\paragraph{Reference chains.} A program contains $c$ chains, each with $d$ assignments, followed by a query and \texttt{Output:}. The root assignment stores a single-token noun; every later assignment names the preceding variable. Chain length counts assignments, including the root. For example, the following prompt mixes two chains of three lines:
\begin{center}
\vspace{-6pt}
\small\begin{tabular}{ll}
\texttt{K = apple} & \texttt{M = pear}\\
\texttt{B = K} & \texttt{Q = M}\\
\texttt{D = B} & \texttt{R = Q}\\
\multicolumn{2}{c}{\texttt{print(D)\quad Output:}}
\end{tabular}
\end{center}
\vspace{-6pt}
Each assignment occupies a separate prompt line. The answer is \texttt{apple}. Names, nouns, and queried chains are randomized. In level order, assignments are grouped by depth and shuffled within each level. In interleaved order, chains are randomly merged while preserving definition before use. A line's \emph{pointer} is the variable on its right-hand side; its \emph{parent line} defines that variable. Appendix~\ref{app:details} gives the prompt header, vocabulary, training, and evaluation protocols.

\paragraph{Models and scoring.} We evaluate thirteen standard base models from Qwen3, Llama, OLMo-3, and Gemma-3 \citep{grattafiori2024llama3,yang2025qwen3,olmo2025olmo3,gemma2025gemma3}, plus DeepSeek-V4-Flash (292B MoE). The looped models are Ouro-1.4B and Ouro-2.6B, trained with four loops of 24 and 48 layers, and Huginn-0125, whose four-layer recurrent core was trained with 32 recurrences on average \citep{geiping2025huginn,zhu2025ouro}. Huginn's initial random state is held fixed per prompt across interventions.

Choice accuracy selects among the chains' root values, with chance $1/c$; exact accuracy requires the correct root to rank first over the full vocabulary. \emph{Reach} is the longest chain followed with at least 80\% accuracy, linearly interpolated at the first downward crossing. If all tested lengths pass, we report a lower bound. Standard-model surveys use three-chain choice accuracy; standard-model LoRA evaluations use two-chain exact accuracy; looped-model experiments use two-chain choice accuracy unless stated. Headline standard results use 200 programs per cell; other sample sizes and uncertainty estimates are specified with their results in the appendix.

\paragraph{LoRA at one layer.} Our rank-8 LoRA \citep{hu2022lora} acts on the residual stream at a layer's input, updating an identity weight with a learned scale:
\begin{equation}
h\leftarrow M(h)=s h+BAh,
\label{eq:LoRA}
\end{equation}
where $n$ is hidden width, $A\in\R^{8\times n}$, $B\in\R^{n\times8}$, and $s$ is a scalar, with no bias. The implementation's RMS factors cancel. In ReFT's terms, this is the DiReFT form of a representation intervention \citep{wu2024reft}. It acts independently at each token; frozen attention and MLP layers carry all communication. Layer indices start at zero. We call attention/MLP projection-weight updates \emph{projection LoRA}. Looped models use the same LoRA in every loop unless specified.

The scalar $s$ rescales the full state; $BA$ is rank eight. Standard Qwen and Ouro LoRAs learn $s=1.0006$ and $1.004$; their longer-trained versions learn $1.008$ and $1.002$. These four scales stay within 1\% of identity (all scales in Appendix Table~\ref{tab:scales}). Only $A$, $B$, and $s$ are trained: 65,537 parameters in Qwen3-8B, 32,769 in Ouro-1.4B, and 84,481 in Huginn. Training combines answer cross-entropy with $\mathrm{KL}(p_0\Vert p_M)$ on WikiText-103 \citep{merity2017wikitext}. The standard Qwen LoRA enters layer 14 and trains through 20 lines; Ouro's enters layer 6 and trains with four loops. Longer-trained Qwen and Ouro LoRAs see up to 40 lines and a larger vocabulary; Huginn's sees up to 24. Huginn's initial 12-line LoRA omits the text penalty. WikiText perplexity changes from 10.14 to 10.148 in Qwen and 13.294 to 13.302 in Ouro at four loops: a measured control, not a general-behavior guarantee.

\begin{figure*}[t]
\centering
\includegraphics[width=\textwidth]{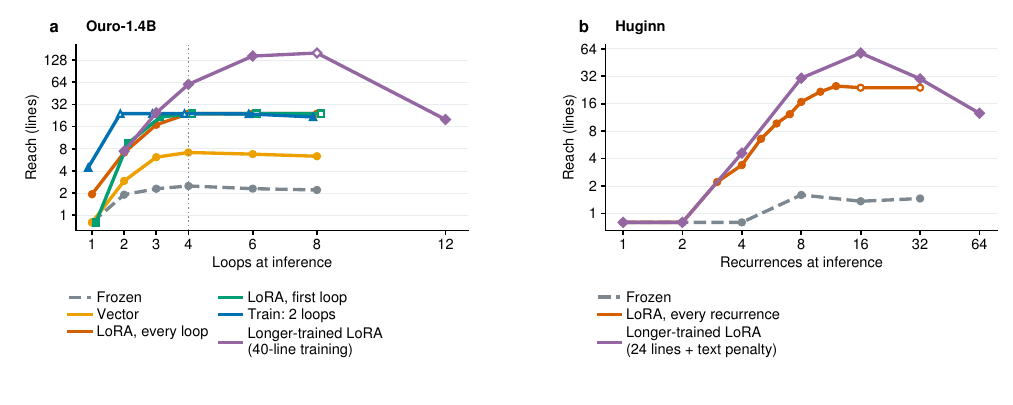}
\caption{\textbf{Extra loops extend the chain once a LoRA starts the relay.} Reach uses two-chain choice accuracy; open markers denote lower bounds. (a) Ouro-1.4B: frozen, a constant vector, and rank-8 LoRAs with different training lengths, training loop counts, or application only in the first loop. The longer-trained LoRA is evaluated through 160 lines; the dotted line marks four training loops. (b) Huginn: frozen and LoRAs trained through 12 or 24 lines; the longer-trained LoRA includes the text penalty. More loops help over a substantial range, but gains eventually reverse.}
\label{fig:reach}
\end{figure*}

\section{The default computation stops early}\label{sec:default}
\paragraph{More layers do not mean more references.} Reliable reach across the thirteen standard models is 1.4--3.6 lines, with median 2.2. Qwen3-8B drops from 83.5\% choice accuracy at three lines to 54\% at four and 41.5\% at five. OLMo-3-32B and OLMo-3-7B both reach about 2.6 lines despite having 64 and 32 layers. DeepSeek-V4-Flash reaches 4.0 lines and scores 42\% at six and 38\% at eight, near the 33\% chance level. Tables and complete traces appear in Appendix~\ref{app:window}.

Additional loops give similarly diminishing returns. Ouro-1.4B's three-chain reach is 0, 1.6, 2.3, 2.2, and 2.2 lines after one through five loops. Ouro-2.6B follows the same pattern despite twice the layers per loop. Huginn reaches only 1.4--1.6 lines after 8, 16, or 32 recurrences. Solved demonstrations improve answer format but leave reach short. The bottleneck therefore persists beyond both additional depth and additional repetitions.

\paragraph{The query reads pointers, then copies the value.} Figure~\ref{fig:default_main}a,b summarizes the computation. We locate the computation by changing one root noun or redirecting one pointer, then restoring a clean residual state into the counterfactual run \citep{meng2022rome}. The recovered fraction of the answer-logit difference measures the restored state's causal effect (Appendix Fig.~\ref{fig:default_std}c,d). In Qwen3-8B, a root value remains at its source until the query takes over near layer 32 of 36, almost regardless of whether the chain has one or four lines. Pointer effects reach the query earlier, with the nearest pointer first. Blocking query attention to pointer lines harms answers in the middle layers but changes accuracy by at most two points between those layers and the late value copy.

The interval over which a pointer's own effect falls from 90\% to 10\% spans $7.1\pm1.4$ layers across models with 16--64 layers. Its width changes little with depth, though individual widths depend on the tracing setup (Appendix~\ref{app:window}). In Ouro, the query takes over pointers chiefly in the third and fourth loops, within a similarly short interval in each loop. The network continues computing after this interval, but it stops extending the chain.

\paragraph{The context makes little progress on its own.} A linear read-out predicts which root a line ultimately refers to from its pointer state. We count a line as readable at 75\% accuracy on two chains and track the longest prefix in which every line passes this threshold. This is our measure of the relay. In frozen Qwen3-8B it reaches line four and stops; across the thirteen models it stops at lines three to five. Ouro never reliably reaches line five, even after six loops, and Huginn never reaches line four after sixteen recurrences. The query can resolve another pointer or two, but neither the program nor the query traverses a long chain.

\begin{figure*}[t]
\centering
\includegraphics[width=\textwidth]{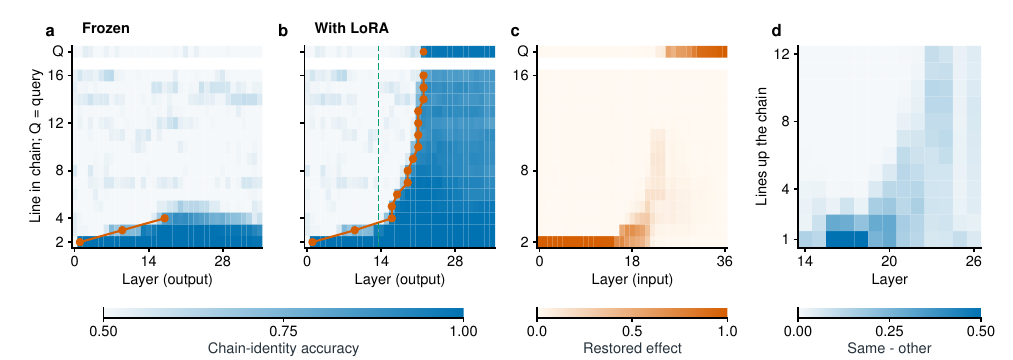}
\caption{\textbf{The LoRA changes how the program's own lines carry the chain.} Qwen3-8B on sixteen-line chains. (a,b) Linear read-outs of chain identity at each pointer and the query, frozen and with the layer-14 LoRA; the marked contour is the first 75\% crossing. (c) Restoring the second line's pointer affects progressively later lines, then the query. Restoration positions are layer inputs. (d) In interleaved programs, attention reaches further up the same chain through the middle layers (same-chain minus other-chain attention). Heads are chosen and scored on separate program halves.}
\label{fig:relay_std}
\end{figure*}

\section{A small LoRA makes the layers count}\label{sec:switch}
\paragraph{Long chains in one pass.} With the layer-14 LoRA, Qwen3-8B answers 16-, 20-, and 24-line chains with 98.5\%, 98.0\%, and 99.0\% exact accuracy, against 15.5\%, 14.5\%, and 15.5\% frozen. The 24-line result is 198 of 200 programs (95\% Wilson interval 96.4--99.7\%); three training seeds score 94.7--97.5\% on separate samples. A longer-trained LoRA answers 40-line chains at 98\% and 48-line chains at 88\%, versus 4\% and 9\% frozen, reaching 50 lines. The longest results use level-ordered, two-chain programs. Interleaving lowers 24-line accuracy to 76.5\%, and three chains lower 16-line accuracy to 58.0\%; both remain well above the frozen model. A separately trained three-chain LoRA transfers from code to English assignments (Appendix~\ref{app:accuracy}).

\paragraph{Longer chains with more loops.} Ouro's standard LoRA reaches 1.9, 7.1, and 17 lines after one, two, and three loops, and answers every tested length through 24 after four. Its longer-trained LoRA reaches 60 after four loops, about 146 after six, and at least 160 after eight (Fig.~\ref{fig:reach}); 160-line accuracy is 87\%, at four times the longest training chain. A second training seed reaches 52 rather than 60 lines after four loops. Huginn's longer-trained LoRA reaches 31 lines after eight recurrences and 58 after sixteen (95\% interval 37--64). These gains are not unlimited: both models lose reach when run well beyond their training loop counts.

Ouro's first-loop-only LoRA reaches 9.5 lines after two loops, 21.5 after three, and about 25 after four through eight. Unmodified loops therefore continue the computation. This result concerns shorter programs; the LoRAs reaching 60 and 160 lines act in every loop. Training with fewer loops also changes the pace: LoRAs trained with one, two, or four loops reach 12.5, 4.4, or 1.9 lines in the first loop. Appendix~\ref{app:accuracy} gives seed, chain-count, vocabulary, and loop-count controls.

\paragraph{The useful layers can be run again.} We also re-enter Qwen3-8B's layers 14--22 and apply the LoRA at each re-entry. With training matched to the longer-chain setting, 64-line exact accuracy rises from 34\% in one pass to 66\% with one re-entry and 92\% with two; the frozen model scores 10\%. Repeating this part of a standard model gives the relay more time to advance.

\paragraph{Other interventions work at the same location.} Rank-8 projection LoRA on layer 14's seven projections yields 99.0\% and 90.0\% accuracy at 16 and 24 lines. FLAS \citep{jin2026flas} is a parallel method that moves the hidden state through a learned flow. Three low-rank integration steps yield 99.5\% and 94.0\%; FLAS-style flow blocks also extend reach. At layer 26, all four interventions select between two chains near chance. This supports a shared placement preference, without establishing the relay mechanism for projection LoRA or flows (Table~\ref{tab:interventions_main}; full controls in Appendix Table~\ref{tab:interventions}).

\begin{table}[!htb]
\centering\small
\caption{\textbf{Different edits share an early placement preference.} Qwen3-8B exact accuracy (\%) on two-chain, 24-line programs; 200 programs per cell with matched training. This comparison evaluation is separate from the headline LoRA result. Step counts are the same at both layers.}
\label{tab:interventions_main}
\begin{tabular}{@{}lccc@{}}
\toprule
Intervention & Parameters & Layer 14 & Layer 26 \\
\midrule
LoRA & 66K & 96.5 & 54.5 \\
Projection LoRA & 606K & 90.0 & 52.5 \\
FLAS flow, 3 steps & 66K & 94.0 & 54.0 \\
FLAS block, 1 step & 168M & 95.5 & 56.0 \\
\bottomrule
\end{tabular}

\end{table}

\begin{figure*}[t]
\centering
\includegraphics[width=\textwidth]{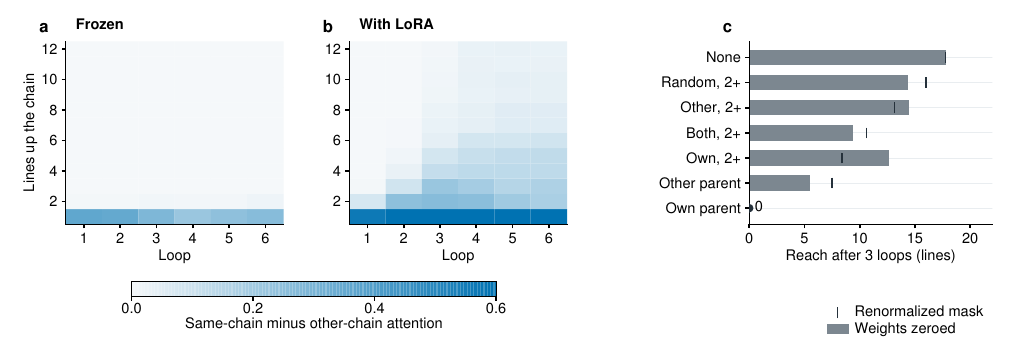}
\caption{\textbf{Longer reads use frozen attention, and the relay needs its parent links.} Ouro-1.4B: (a,b) attention to earlier pointers on the same chain minus the other chain, frozen and with LoRA, on 24-line programs; heads are selected and scored on separate program halves, then maximized over each loop's layers. (c) Two-chain reach after three loops when program lines cannot attend to the specified assignments in any layer or loop; query attention is unchanged. ``2+'' means two or more lines up; Own/Other/Both identify chains. Bars zero weights after softmax; ticks renormalize. Parent cuts stop the relay, while longer-distance cuts reduce its reach. Full head and name-decoding evidence is in Appendix~\ref{app:mechanism}.}
\label{fig:mechanism_main}
\end{figure*}

\section{The LoRA starts a relay in the middle layers}\label{sec:mechanism}

\paragraph{A short interval carries a much longer computation.} With the LoRA, Qwen3-8B's relay reaches lines 5, 6, 6, 8, 9, 13, and 16 at the outputs of layers 16--22 (Fig.~\ref{fig:relay_std}). The frozen model reaches line four at layer 17 and goes no further. Restoring a redirected pointer's clean state recovers its effect at progressively later program lines and then at the query, which recovers half the effect from layer 25. Thus the gain is visible in both read-outs and causal interventions.

The same pattern repeats across loops. Ouro's relay advances almost entirely within layers 7--15 of each loop; with the longer-trained LoRA it reaches lines 4, 9, 25, and 40 after the first four loops (Fig.~\ref{fig:default_main}d). The deeper Ouro-2.6B still uses a short middle interval, and Huginn advances mainly in the last two layers of its recurrent core. The LoRA's location can change when this relay starts, but the frozen layers carry it.

\paragraph{Attention reaches further as the relay advances.} We measure attention from a line's pointer to pointers earlier in its own chain, subtracting attention to corresponding lines of the other chain. A head is selected on one half of the programs and scored on the other. Without the LoRA, only attention to the parent line consistently distinguishes the chains. With the LoRA, Qwen's attention reaches two lines up at layers 16--18, then three, four, six, and seven at layers 19--22. Each layer advances the relay by at most that distance. In Ouro, attention reaches one, two, four, and five lines up across four loops. This gradual growth differs from pointer doubling: attention overlaps several earlier lines rather than making discrete jumps that double in length.

The relevant heads already exist in the frozen model. In Ouro, heads that previously read no further than the parent line take over longer reads, while the strongest frozen parent-line heads themselves read less far. At four loops, ablating seven longer-reading heads lowers reach to 12.0, versus median 24 across twenty layer-matched random-head draws. In Huginn, the strongest parent-reading head directly takes over longer reads (Appendix~\ref{app:mechanism}). In Qwen3-8B on interleaved sixteen-line chains, removing ten parent-reading heads lowers choice accuracy from 89\% to 53\%, versus 83\% after removing ten random heads matched by layer. Removing eight longer-reading heads leaves 65\%, versus 90\% for eight matched random heads. These cuts test the contribution of the heads identified by their reads.

\begin{figure*}[t]
\centering
\includegraphics[width=\textwidth]{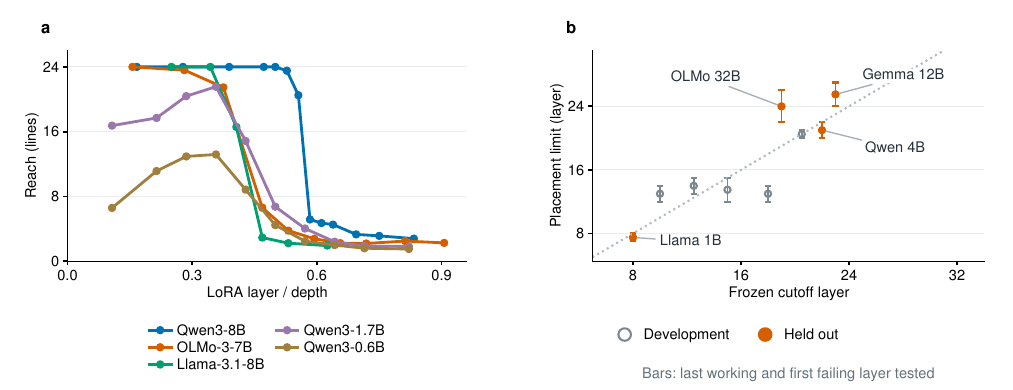}
\caption{\textbf{A model-specific layer separates effective and ineffective placements.} (a) Reach of independently trained rank-8 LoRAs versus input layer relative to model depth; exact accuracy on two-chain programs, capped at 24 lines. (b) The frozen-model cutoff versus the measured placement limit in nine models. Bars bracket the last working layer and next tested layer; filled markers identify held-out models. The diagonal denotes agreement.}
\label{fig:placement}
\end{figure*}

\paragraph{The parent line is necessary.} Removing each Qwen pointer's attention to its parent in layers 14--22 returns six-, eight-, and twelve-line chains to chance (53\%, 48\%, and 55\%). Removing the same edges after the relay, in layers 23--29, leaves 100\%, 100\%, and 98\%. In Ouro, removing parent attention stops the relay; removing attention two or more lines up reduces reach from 17.8 to 9.4 (Fig.~\ref{fig:mechanism_main}). On three-chain Ouro programs, cuts to the line's own chain matter, while matched cuts to other chains or random lines have little effect. In Ouro, parent attention remains necessary with weights zeroed after softmax and with renormalizing masks (Appendix~\ref{app:masks}).

\paragraph{Earlier names let a line find earlier lines.} Linear probes decode names defined progressively further up the chain after the LoRA. In Ouro, names three to six lines up are present even at lines that the relay has not yet reached. Blocking parent-line attention lowers loop-two decoding of names four lines up from 0.29 to 0.18, near loop one's 0.13; once a line's chain is resolved, the names fade. The observations support a computation in which parent attention brings names down the chain, and longer attention uses them to find the chain's earlier lines.

The LoRA itself exchanges no information across tokens. Applying it only to program tokens retains the gain, while applying it only at the query adds just two to four lines. Chain identity later occupies the broader residual stream, not just the low-rank part's eight-dimensional span: removing that span from later states leaves read-out near 95\%. Ouro's unmodified loops continue the relay after a first-loop-only LoRA. Together, these results place the extended computation in the frozen layers.

\paragraph{The ability is learned before the LoRA.} The same LoRA-training procedure unlocks progressively longer chains in later OLMo pretraining checkpoints: reach is 10.2 after 84 billion tokens and 23.7 after 336 billion, while frozen reach stays below two. Small looped transformers trained from scratch on the programs learn the same relation between relay progress and attention distance, matching exactly in 108 of 165 layer steps, advancing further in 8 and less far in 49. Both findings connect the intervention to an available learned computation (Appendices~\ref{app:transfer} and~\ref{app:toy}).

\begin{figure*}[t]
\centering
\includegraphics[width=\textwidth]{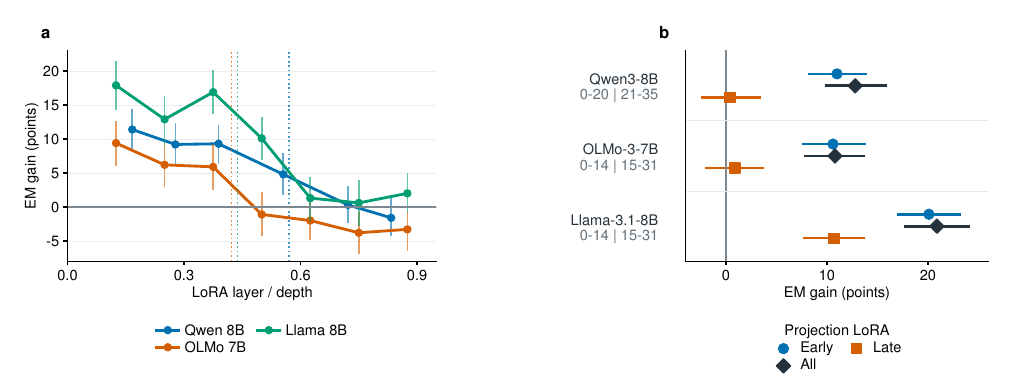}
\caption{\textbf{Early adaptation preserves most of the MuSiQue improvement.} Results use identical prompts on 900 development questions with gold paragraphs. (a) Exact-match gain from LoRAs trained on MuSiQue, versus layer relative to depth; dotted lines mark placement limits measured on programs. (b) Rank-8 projection LoRA restricted to early, all, or late layers. Error bars are paired 95\% bootstrap intervals over questions. Projection LoRA learning rates are $3\cdot10^{-4}$ for Qwen and $10^{-4}$ for OLMo and Llama.}
\label{fig:qa_std}
\end{figure*}

\section{Where to put the fix}\label{sec:placement}

\paragraph{A sharp placement limit.} Holding LoRA rank, data, and optimization fixed, moving the Qwen3-8B LoRA from layer 20 to 21 lowers reach from 20.5 to 5.2 lines (Fig.~\ref{fig:placement}). Corresponding transitions occur at layers 12--15 in OLMo-3-7B and 13--15 in Llama-3.1-8B, and second seeds reproduce them. The limit lies within the middle layers: an early-started relay can continue beyond it, but a later LoRA cannot start one. In Qwen, the layer-20 LoRA enables increasingly long attention reads; the layer-21 LoRA leaves attention near the parent line.

\paragraph{Predicting placement from the frozen model.} We define the \emph{cutoff layer} from frozen three-line, three-chain programs: the first layer where restoring a pointer's own token recovers less than half its effect, averaged over pointers on lines two and three. Qwen's cutoff is 20.5. A placement works if it recovers at least half the best observed extension above median frozen reach; the last working placement and next tested layer bracket the limit.

Before obtaining held-out results, we preregistered the cutoff prediction and two alternatives fitted or chosen using five development models. The criterion was agreement within the observed bracket expanded by one layer, in at least three of four held-out models. The cutoff met this criterion for Llama-3.2-1B, Qwen3-4B, and Gemma-3-12B, but missed OLMo-3-32B by five layers from the bracket midpoint (Table~\ref{tab:heldout_main}). Its mean absolute error was 2.25 layers, versus 3.45 for 45\% of depth and 3.37 for the value-copy layer minus 11.5. Across all nine models, however, its error is 2.22 layers, essentially tied with 2.27 for a post-hoc rule at 48\% of depth. Thus the cutoff provides a starting point for a local sweep, without demonstrated superiority to relative depth (Appendix~\ref{app:placement}).

\begin{table}[!htb]
\centering\small
\caption{\textbf{The frozen cutoff locates three held-out limits.} Brackets join the last working and next tested layer. Checks allow one layer beyond either edge, as preregistered. All entries are layer indices; all nine models and baseline rules appear in Appendix Table~\ref{tab:edges}.}
\label{tab:heldout_main}
\begin{tabular}{@{}lcc@{}}
\toprule
Held-out model & Placement bracket & Cutoff \\
\midrule
Llama-3.2-1B & 7--8 & 8.0\rlap{$^\checkmark$} \\
Qwen3-4B & 20--22 & 22.0\rlap{$^\checkmark$} \\
Gemma-3-12B & 24--27 & 23.0\rlap{$^\checkmark$} \\
OLMo-3-32B & 22--26 & 19.0 \\
\bottomrule
\end{tabular}

\end{table}

\paragraph{A later loop provides another opportunity.} Ouro's layer-20 LoRA follows one loop's middle layers but precedes the next loop's. It therefore improves reach with a delay. Applying a LoRA only in the last loop isolates the distinction: layer 6 reaches 21.6 lines after four loops; layer 20 reaches 2.6, essentially the frozen 2.5. Ouro-2.6B separates this from unused depth: a last-loop-only LoRA at layer 24 still has 24 layers after it, but reaches just 4.7 lines after four loops versus 2.6 frozen. The same application schedule at layer 12, inside the middle range, reaches 9.6. What matters is whether useful middle-layer computation can still follow the intervention, not how many layers remain in total.

\section{Multi-hop question answering}\label{sec:downstream}

\paragraph{Fictional facts preserve the controlled setting.} We replace assignments with typed facts about fictional entities and ask questions that compose the relations. A Qwen3-8B LoRA trained on two- to five-hop questions raises exact match from 41.2\% to 97.4\% over 500 paired questions, a gain of 56.2 points [51.8, 60.6]. It also improves unseen six-hop questions from 23\% to 88\%. Ouro's LoRA, trained on one to four hops, extends reliable answers from two hops to six after four loops and eight after six or eight loops, at 91--92\% accuracy, twice the training hops.

\paragraph{Early layers also matter on MuSiQue.} We train both LoRA forms separately on MuSiQue questions, supervising answers only, and evaluate direct answers with shuffled supporting paragraphs. This tests placement on a benchmark, without transferring a program-trained LoRA. On 900 development questions, LoRAs at the earliest tested layers add 11.4, 9.4, and 17.9 exact-match points in Qwen3-8B, OLMo-3-7B, and Llama-3.1-8B (Fig.~\ref{fig:qa_std}). Qwen's frozen exact match is 52.9\%; the layer-6 LoRA beats the layer-30 LoRA by 13.0 points [10.1, 16.0]. These are also the best tested layers on the same development set, with no held-out layer-selection split. At the fixed program intervention layer 14, Qwen gains 9.3 points averaged over three seeds. Late LoRAs add little, with Llama's useful region extending beyond its program-based limit.

Early projection LoRA retains most of its all-layer gain: 11.0 versus 12.8 points in Qwen, 10.6 versus 10.8 in OLMo, and 20.1 versus 20.9 in Llama. Late projection LoRA gains 0.4, 0.9, and 10.7 points, respectively. Equal-width layer ranges in Qwen show that parameter count does not explain the placement effect. At layer 6, a FLAS-style low-rank flow and flow block gain 9.4 and 9.3 points. Full results, seed variation, intervals, and training details appear in Appendix~\ref{app:musique}.

In Ouro, the layer-6 LoRA improves exact match on 300 development questions from 48.3\% to 60.0\% after four loops, with the largest gain on four-hop questions (40\% to 58\%). Gains persist from two through eight loops. With both LoRAs applied every loop, layer 20 gains 7.0 points at four loops, versus layer 6's 11.7 (Fig.~\ref{fig:teaser}c). Projection LoRA of comparable size at the same layer also improves answers. Last-loop-only LoRAs again distinguish location: layer 6 adds 6.3 points after two, four, or six loops, whereas layer 20 adds 2.0, $-1.3$, and $-1.3$. The benchmark supports the placement preference; the program experiments establish the relay mechanism.

\section{Related work}\label{sec:related}
\paragraph{Composition and information flow.} The gap between knowing facts and composing them motivates direct-answer multi-hop studies \citep{press2023compositionality,yang2024latentmultihop,balesni2024twohop}. Layer ordering and competing context/query dependencies can obstruct composition \citep{biran2024hopping,lepori2025racing,yang2025internalcot}; back-attention and intermediate-layer adaptation address related access problems \citep{sia2024icl,yu2025backattention}. Two-hop tasks expose layerwise computation, while depth alone need not improve composition \citep{guo2025twohop,csordas2025depth}. We connect these timing questions to the computation accessible in frozen pretrained weights and a prospective test of intervention placement.

\paragraph{Binding and reusable circuits.} Variable assignments probe composition \citep{zhang2022lego,hsieh2024ruler}, alongside entity-tracking tasks \citep{kim2023entity,tang2026entitystate}. Binding representations, lookbacks, and entity-tracking circuits explain how models preserve references \citep{feng2024binding,prakash2024entity,prakash2025lookback,wu2025binding,oh2026rebinding}. Mechanistic analysis of a transformer trained on symbolic multi-step reasoning provides a close precedent \citep{brinkmann2024symbolic}. Here, the central contrast is between the short default computation of pretrained models and the longer relay enabled by a tiny edit. Parent-reading heads resemble induction heads \citep{olsson2022induction}; growing sets of earlier lines relate to learned graph-search computations \citep{saparov2025search}. We use causal tracing and attention knockout methods \citep{meng2022rome,wang2023ioi,geva2023dissecting} to distinguish these reads from correlations in probes.

\paragraph{Loops and adaptation.} Recurrent-depth models make inference computation adjustable \citep{dehghani2019universal,giannou2023looped,yang2024looped,saunshi2025latent,fan2025looped}. Pretrained looped models and retrofits realize this possibility \citep{geiping2025huginn,zhu2025ouro,mcleish2025retrofit,bae2025mor,koishekenov2025etd}, but their repeated stages need not produce a latent reasoning chain \citep{lu2025latentcot,blayney2026stages,wang2026jlensloop}. Unlike training recurrent models for $k$-hop extrapolation \citep{kohli2026loop} or supervising one hop per loop \citep{shapiro2026}, we start the relay in pretrained loop bodies with a representation edit. We compare the residual-stream LoRA with projection LoRA \citep{hu2022lora} and FLAS representation interventions \citep{jin2026flas}. Localized adaptation also has precedents in LoFiT \citep{yin2024lofit} and single-layer reinforcement learning near mid-depth \citep{zhang2026onelayer}. We use established intervention forms to measure where frozen layers can extend an in-context computation; Appendix~\ref{app:related} gives further connections.

\section{Discussion}\label{sec:discussion}
A wrong direct answer or weak scaling with loops does not establish that the weights cannot compose the references. The default pass combines a short context relay, a few query-side steps, and a late value copy; the LoRA lets the relay continue while the middle layers can act. Evaluation should therefore distinguish default behavior from computation accessible under a specified intervention. For adaptation, placement determines which useful computation can follow a change; for inference, loops and re-entry help by repeating the layers that carry it.

\paragraph{Limitations.} The mechanistic tasks are synthetic and in-context; MuSiQue provides evidence about placement, with gold paragraphs in the main setting, rather than direct evidence of the same mechanism. Detailed traces focus on Qwen3-8B and Ouro-1.4B, and the looped models cover two families, with Ouro-2.6B grown from Ouro-1.4B. LoRAs are trained for each task format. The text penalty is a measured control, placement predictions have modest precision, and first-loop-only sufficiency is established only through about 25 lines. The longest-chain looped results use level order and a LoRA in every loop. The learned routing directions and minimum sufficient rank remain unidentified; our causal tests constrain the computation without uniquely specifying its algorithm.

\label{mainbodyend}
\clearpage
\restorePaperLayout
\section*{Impact Statement}
This work diagnoses how pretrained language models follow references supplied in context. It uses public models, synthetic tasks, and public datasets. The interventions are evaluated on specific task families; deployment reliability and effects on broader model behavior require separate evaluation. We do not foresee direct harmful uses beyond those of the underlying models.
\begingroup
\let\url\nolinkurl
\bibliography{refs}
\endgroup
\bibliographystyle{icml2026}
\clearpage
\appendix
\onecolumn
\raggedbottom
\clubpenalty=10000
\widowpenalty=10000
\displaywidowpenalty=10000
\section{Tasks, interventions, and measurements}\label{app:details}

\subsection{Prompts, models, and scoring}
A program starts with \texttt{Here is a short program. Each line assigns a value to a variable.}, lists one assignment per line, and ends with \texttt{print(X)} and \texttt{Output:}. Each chain begins with a single-token noun and continues with assignments to previously defined variables. Length counts assignments in one chain, including its root. Level order shuffles the chains within each level; interleaved order randomly merges the chains while preserving their individual order. Names, root values, and the queried chain are drawn at random. The standard-model panel uses 26 capital letters. Longer programs add lowercase letters and tokenizer-compatible two-letter names; the full pools contain 369 names for Ouro and 527 for Huginn.

DeepSeek-V4-Flash uses the DeepSeek-V4-Flash-Base checkpoint. The standard panel comprises thirteen base models: Qwen3 from 0.6B to 14B, Llama-3.2-1B and 3B, Llama-3.1-8B, OLMo-3-7B and 32B, and Gemma-3-4B, 12B, and 27B. Ouro-1.4B and 2.6B apply 24 and 48 layers per loop and were trained with four loops. We change the loop count at inference. Huginn-0125 has two embedding layers, a four-layer recurrent core, and two decoding layers. Each recurrence receives the embedded prompt; its latent state is initialized randomly, with the same initial state used for matched runs on a prompt. Huginn was pretrained with a random recurrence count averaging 32. Layer indices begin at zero. An intervention at layer $a$ acts on that layer's input, after $a$ layers; read-outs and attention refer to layer outputs.

Choice accuracy compares logits only among the chains' root values, with chance $1/c$ for $c$ chains. Exact accuracy requires the correct root to have the highest logit in the vocabulary. Reach is the first downward crossing of 80\% accuracy, linearly interpolated between tested lengths; if every tested length exceeds the threshold, the maximum tested length is a lower bound. The standard frozen panel uses three chains and choice accuracy, standard-model LoRAs use two chains and exact accuracy, and looped-model evaluations use two chains and choice accuracy unless specified otherwise.

Panel cells and headline standard-model evaluations contain 200 programs; placement sweeps and looped evaluations ordinarily contain 150. Longer-trained LoRAs, lengths above 24, and query-blocking, few-shot, mask, and head-ablation runs generally use 100; Huginn, longer-trained LoRA masks, and the longer-trained LoRA's 128- and 160-line cells use 60. Individual experiments below state exceptions, including the Qwen3 head comparison with 150 programs and the larger mask controls.

Fictional-fact prompts contain an instruction, a solved example, and facts about fictional people, places, works, and organizations, with distractor chains sharing the same relations. Their compositional questions are scored by the first token of each chain's final entity. MuSiQue outputs are decoded greedily, stopping at a newline or after 16 tokens in standard models and 12 in Ouro. Exact match and F1 use the answer and its aliases after lowercasing and removing punctuation and articles.

\subsection{Training the LoRA}
The LoRA in Section~\ref{sec:setup} is implemented as
\begin{equation}
 h\leftarrow s h+\operatorname{rms}(h)BA\frac{h}{\operatorname{rms}(h)}=(sI+BA)h.
\end{equation}
It acts independently at each token; frozen attention and MLP layers perform all communication between tokens. At rank eight, the learned parameter counts are 65,537 for Qwen3-8B, 32,769 for Ouro-1.4B, and 84,481 for Huginn. Ouro normally applies the same LoRA in every loop at layer 6; Huginn applies it at the recurrent core's input-adapter output. The standard Qwen3-8B LoRA enters layer 14.

Table~\ref{tab:scales} reports the learned scalar in each LoRA. The standard and longer-trained Qwen3-8B LoRAs and the every-loop Ouro LoRAs keep $s$ within 1\% of one; the first-loop-only, Huginn, and re-entry LoRAs learn larger rescalings. The scalar acts on the full residual stream, while $BA$ is the rank-8 part of the intervention.

\begin{table}[!htb]\centering\small
\caption{\textbf{Learned scales distinguish the LoRA's scalar and low-rank components.} Each value is the learned $s$ in $M(h)=(sI+BA)h$. Longer-trained LoRAs use the training lengths described below; the Qwen3-8B re-entry LoRA accompanies additional passes through layers 14--22.}\label{tab:scales}
\begin{tabular}{lr}
\toprule
LoRA & Learned scale $s$ \\
\midrule
Qwen3-8B, layer 14 & 1.0006 \\
Qwen3-8B, longer-trained LoRA & 1.008 \\
Ouro-1.4B, layer 6, every loop & 1.004 \\
Ouro-1.4B, longer-trained LoRA & 1.002 \\
Ouro-1.4B, first loop only & 1.039 \\
Huginn, trained up to 12 lines & 1.06 \\
Huginn, longer-trained LoRA & 1.09 \\
Qwen3-8B, re-running layers 14--22 & 0.858 \\
\bottomrule
\end{tabular}

\end{table}

For standard models, initialize $A$ with independent normal entries of standard deviation $n^{-1/2}$, $B=0$, and $s=1$. AdamW uses learning rate $10^{-3}$, no weight decay, and gradient clipping at one. Training lasts 1,200 steps with batches of 16 two-chain, level-order programs, with lengths sampled up to 20. The long variant uses 2,000 steps, lengths up to 40, and the larger name vocabulary. Each step adds $\mathrm{KL}(p_0\Vert p_M)$ on eight 160-token WikiText-103 training passages, weighted by one, to the answer cross-entropy. Placement sweeps use this same recipe.

The text control evaluates the first 64 WikiText-103 test passages longer than 800 characters, truncated to 256 tokens. Qwen3-8B's loss changes from 2.3165 to 2.3173 and perplexity from 10.14 to 10.148. The 24-line headline evaluates 198 correct answers among 200 fresh programs, with 95\% Wilson interval 96.4--99.7\%. On a separate sample, the same training seed scores 195/200; two further seeds score 142/150 and 143/150. The 48-line result is 88/100, with interval 80.2--93.0\%.

Ouro program LoRAs use 1,200 steps, batches of 16, and lengths drawn uniformly from 1 to 20, or 1 to 40 for the longer-trained LoRA. Gradients pass through all four training loops. The text penalty uses eight WikiText chunks of 160 tokens per step with weight one; at four loops, perplexity changes from 13.294 to 13.302. Fact training uses eight questions of one to four hops per step. MuSiQue uses 1,500 steps with eight questions, their supporting paragraphs and optionally two distractor paragraphs, answer-token loss, and four WikiText chunks per step.

Huginn training uses batch size one with eight accumulated programs per update and gradients through eight recurrences. The initial latent-state seed is fixed per program. The first LoRA uses 600 steps, chains of 1--12 lines, and no text penalty. The longer-trained LoRA uses 1,000 steps; its maximum chain length increases from 8 to 24 over the first 500. One WikiText chunk is used every four programs. At sixteen recurrences, perplexity is 15.7 frozen and 16.0 with this LoRA.

\subsection{Restoration, read-outs, and attention}
Let $g(x)=z_v(x)-z_{v'}(x)$ compare final-token logits for the clean and counterfactual answers. Replacing only the residual state at position $p$ and layer input $\ell$ gives the normalized restoration effect
\begin{equation}\label{eq:restoration}
 S(p,\ell)=\frac{g(x_{\rm cf};h^{\rm clean}_{p,\ell})-g(x_{\rm cf})}{g(x_{\rm clean})-g(x_{\rm cf})}.
\end{equation}
Effects may be negative or exceed one and do not sum to a conserved total across positions. Root-value interventions replace a noun while preserving its chain. Pointer interventions redirect one assignment to another chain. Detailed traces retain pairs answered correctly in both versions, so the five-line traces describe successful computation at that length. The panel's cutoff uses 24 retained pairs per pointer position on three-line programs. For the line-2 pointer, only 13 pairs in Llama-3.2-1B and 11 in Qwen3-0.6B survive 1,440 attempts.

For each line and layer output, a ridge read-out predicts chain membership from the pointer state. The 600 programs split into 450 training and 150 test examples; the ridge penalty is 0.1 times the mean squared state norm. A line is readable at 75\% accuracy with two chains, or 60\% against 33\% chance with three. The relay reaches line $k$ only when every line through $k$ is readable. Chain membership depends on pointers but not on the root noun, separating the computation of where to read from copying the value itself.

Attention $j$ lines up the chain is attention from a line's pointer to the pointer $j$ assignments earlier on its own chain, minus attention to the corresponding line of the other chain. Heads for each layer or loop and distance are selected on half the programs and evaluated on the other half. The Qwen3-8B parent-line cut blocks the four tokens of the parent assignment from each pointer and renormalizes; it uses 100 programs per length. Head ablations remove only heads at or after the LoRA: ten parent-reading heads in layers 14--22 and 34, or eight heads reading further in layers 17--24. Their random controls match the number removed at each layer and use 150 programs per point.

\section{How the frozen models stop}\label{app:window}

\subsection{A short computation in standard models}
Reliable reach ranges from 1.4 to 3.6 lines, with median 2.2, in the panel (Table~\ref{tab:panel}). Qwen3-8B's choice accuracy is 83.5\%, 54\%, and 41.5\% at three, four, and five lines, giving reach 3.1. Doubling OLMo-3's depth from 32 to 64 layers leaves reach near 2.6. DeepSeek-V4-Flash (292B MoE), with 43 layers, reaches 4.0 on the same likelihood-scored task with 100 programs per length. Its choice accuracy is near chance at longer lengths: 0.42 at six lines and 0.38 at eight, against 0.33 chance. For one- to four-line chains, its correct root first becomes the top candidate in at least half the programs after the same 23 layers.

\begin{figure}[!htb]\centering
\includegraphics[width=\textwidth]{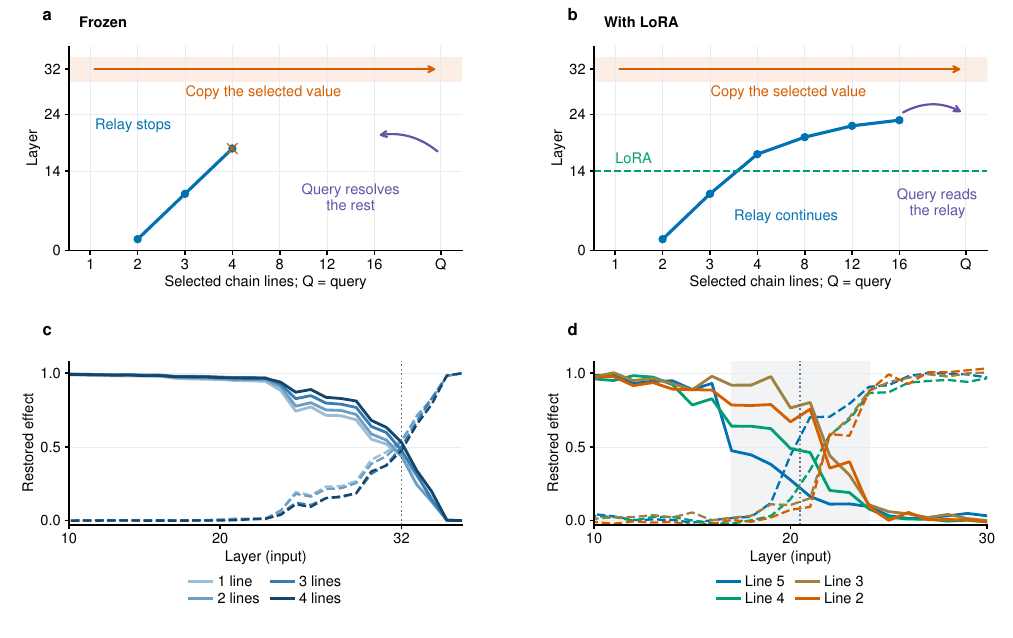}
\caption{\textbf{The query stops following pointers before it copies the answer.} Qwen3-8B: (a,b) the default and LoRA-enabled computations, summarized from separate measurements; (c) root-value restoration at the root token and query, for one- to four-line chains; (d) pointer restoration on correctly answered five-line chains. Solid curves show the source token and dashed curves the query. The nearest pointer reaches the query first. The dotted cutoff is 20.5; value copying occurs near layer 32. Positions denote layer inputs, and restoration curves need not sum to one.}\label{fig:default_std}
\end{figure}

Three measurements locate this failure (Figure~\ref{fig:default_std}). First, Qwen3-8B keeps the root-value effect at its source for most of the pass; the query acquires half near layer 32 of 36, largely independently of chain length. Second, the program resolves only its first assignments: line 2 is readable directly at its pointer, line 3 from layer 9, and line 4 from layer 17; later layers add no further line. Interleaving the chains preserves this short relay. Across the thirteen models, the relay stops at lines 3--5, with line 5 reached only by Qwen3-14B. Third, successful five-line answers involve the query resolving the last one or two pointers, nearest first. The query can therefore meet an incomplete relay: for a query about line 5, following its pointer to a readable line 4 suffices.

\begin{table}[!htb]
\centering\small
\caption{\textbf{Frozen reach stays short across thirteen standard models.} Choice accuracy uses three chains and 200 programs per cell. Cutoff and value-copy layers are relative to depth; value copy is the first layer at which the final token holds half the root-value effect, averaged over one- to three-line chains. Relay is the last consecutively readable line in two-chain, eight-line programs. Unlocked reach uses exact accuracy with a rank-8 LoRA at the best tested layer: six to thirteen placements per model, three for Llama-3.2-3B, and two for Qwen3-14B. A dash denotes an untrained LoRA.}\label{tab:panel}
\resizebox{\textwidth}{!}{\begin{tabular}{lrrrrrrrrrrrr}
\toprule
Model & Layers & \multicolumn{6}{c}{Choice accuracy by chain length (chance $1/3$)} & Reach & Cutoff & Value copy & Relay & Unlocked \\
\cmidrule(lr){3-8}
 & & 1 & 2 & 3 & 4 & 5 & 6 & (lines) & (rel.\ depth) & (rel.\ depth) & (line) & (lines) \\
\midrule
Llama-3.2-1B & 16 & 0.98 & 0.50 & 0.41 & 0.29 & 0.30 & 0.36 & 1.4 & 0.50 & 0.94 & 3 & 6 \\
Qwen3-0.6B & 28 & 0.99 & 0.52 & 0.34 & 0.35 & 0.25 & 0.39 & 1.4 & 0.36 & 0.92 & 3 & 13 \\
Qwen3-1.7B & 28 & 0.99 & 0.76 & 0.40 & 0.35 & 0.27 & 0.32 & 1.8 & 0.64 & 0.92 & 3 & 22 \\
Llama-3.2-3B & 28 & 0.98 & 0.78 & 0.47 & 0.28 & 0.28 & 0.38 & 1.9 & 0.45 & 0.83 & 3 & $\geq$24 \\
Qwen3-4B & 36 & 0.99 & 0.77 & 0.51 & 0.36 & 0.34 & 0.35 & 1.8 & 0.61 & 0.92 & 4 & $\geq$24 \\
Gemma-3-4B & 34 & 1.00 & 0.85 & 0.54 & 0.36 & 0.37 & 0.32 & 2.2 & 0.51 & 0.85 & 3 & -- \\
Llama-3.1-8B & 32 & 0.99 & 0.81 & 0.49 & 0.32 & 0.28 & 0.28 & 2.0 & 0.39 & 0.79 & 4 & $\geq$24 \\
OLMo-3-7B & 32 & 1.00 & 0.95 & 0.69 & 0.38 & 0.41 & 0.36 & 2.6 & 0.47 & 0.77 & 3 & $\geq$24 \\
Qwen3-8B & 36 & 1.00 & 0.98 & 0.83 & 0.54 & 0.41 & 0.32 & 3.1 & 0.57 & 0.89 & 4 & $\geq$24 \\
Gemma-3-12B & 48 & 0.99 & 0.98 & 0.94 & 0.62 & 0.39 & 0.34 & 3.4 & 0.48 & 0.82 & 4 & $\geq$24 \\
Qwen3-14B & 40 & 1.00 & 0.94 & 0.94 & 0.69 & 0.48 & 0.34 & 3.6 & 0.53 & 0.88 & 5 & $\geq$24 \\
Gemma-3-27B & 62 & 1.00 & 1.00 & 0.96 & 0.69 & 0.46 & 0.36 & 3.6 & 0.48 & 0.87 & 4 & -- \\
OLMo-3-32B & 64 & 1.00 & 0.99 & 0.68 & 0.43 & 0.39 & 0.34 & 2.6 & 0.30 & 0.66 & 4 & $\geq$24 \\
\bottomrule
\end{tabular}
}
\end{table}

\begin{figure}[!htb]\centering
\includegraphics[width=\textwidth]{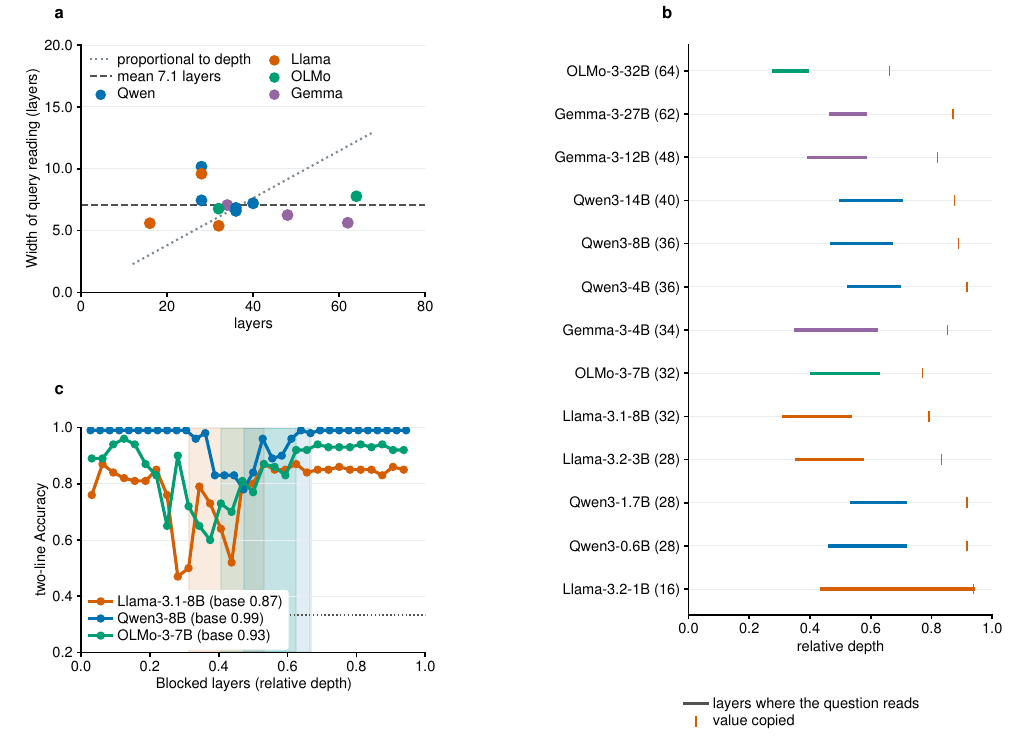}
\caption{\textbf{Pointer reading occupies a similar number of layers at different depths.} (a) Width across thirteen models; dotted: proportional-to-depth width; dashed: mean. (b) Relative positions of pointer reading and value copying. (c) Two-line choice accuracy after blocking query attention to pointer assignments in three consecutive layers, with three chains. Shading marks each model's measured pointer-reading range; the horizontal dotted line marks chance (one third).}\label{fig:window}
\end{figure}

\subsection{Measuring when pointer reading ends}
The cutoff layer is the average, over pointers on lines 2 and 3 of a three-line, three-chain program, of the first layer where restoring the pointer's source token recovers less than half its effect. Qwen3-8B's cutoff is 20.5, the transition is largely complete by 24, and value copying follows near 32.

We measure the full reading range from the first layer where the queried pointer retains less than 90\% of its effect to the first layer where the final token holds 90\% of both pointers' effects. A second width uses only the queried pointer's departure, from 90\% to 10\%, denoted W80. Across 16--64-layer models, W80 is $7.1\pm1.4$ layers (range 5.4--10.2), with slope $-0.017$ against depth; a width proportional to depth would give 0.19. The two-threshold range is $7.2\pm1.1$ layers, with slope 0.011 and bootstrap 95\% interval $[-0.018,0.118]$. Its start scales with depth, at $0.42\pm0.08$ of the stack.

Threshold sensitivity supports interpreting the common width rather than individual-model differences. Seven of nine combinations of start thresholds 0.9, 0.75, 0.5 and end thresholds 0.5, 0.75, 0.9 give slopes between $-0.008$ and 0.075. The other two, starting at 0.5 and ending at 0.5 or 0.75, give 0.116 and 0.111. These measure only the lag between half-departure of the queried pointer and arrival of both pointers: at most two layers in most models up to 36 layers, five in Qwen3-0.6B and OLMo-3-7B, and four to six at depths 40--64. Using two chains shifts individual widths by up to four layers, including 7 to 11 in Qwen3-0.6B and 7 to 4 in Llama-3.1-8B. Within Qwen, Llama, and Gemma, larger models nevertheless resolve more pointers in this similarly wide range.

Query-attention cuts preserve root assignments so that the answer value remains available. Blocking between the end of pointer reading and value copying changes two-line accuracy from 0.99 to 0.98 in Qwen3-8B, 0.93 to 0.94 in OLMo-3-7B, and 0.87 to 0.85 in Llama-3.1-8B. Blocking the reading range gives 0.76, 0.40, and 0.53; blocking all preceding layers gives 0.39, 0.50, and 0.42. Sliding three-layer cuts hurt only at layers 13--22, 6--17, and 7--15 respectively. Attention begins a few layers before the pointer effects leave their source and ends with the measured range.

\subsection{Extra loops repeat the short computation}
\begin{figure}[!htb]\centering
\includegraphics[width=\textwidth]{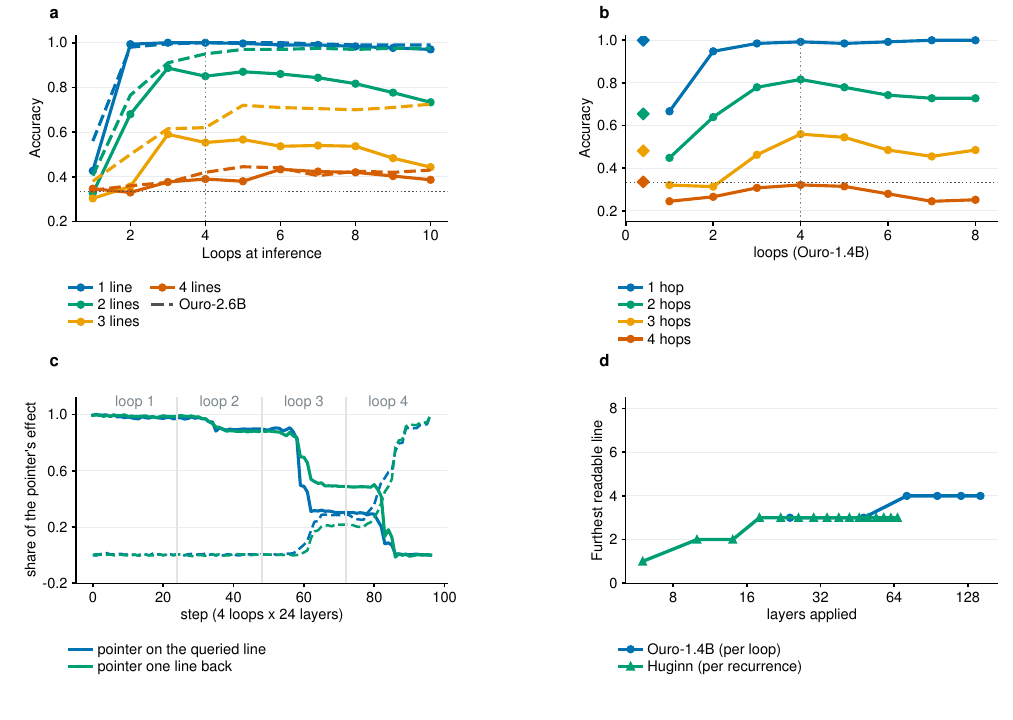}
\caption{\textbf{Frozen looped models gain little after the first few loops.} (a) Choice accuracy on three-chain programs for Ouro-1.4B and 2.6B. (b) Fictional-fact questions; diamonds show Qwen3-1.7B-Base in one pass. (c) A pointer's restoration effect stays at its source (solid) through early loops and reaches the query (dashed) in late loops. (d) Furthest consecutively readable line so far, by layers applied, in Ouro and Huginn.}\label{fig:default_loop}
\end{figure}

Ouro-1.4B answers one-line chains at 43\% after one loop (three chains; chance 33\%). At two loops it answers one-line chains at 99\% and two-line chains at 68\%; at three, two-line chains at 89\% and three-line chains at 59\%. Reach after one through five loops is 0, 1.6, 2.3, 2.2, and 2.2. Four-line accuracy remains within ten points of chance through ten loops. Ouro-2.6B follows a similar schedule despite twice the layers per loop: 0, 1.8, 2.4, 2.5, and 2.7 lines. Thus the second loop adds one to two lines, the third 0.4--0.7 lines or hops, and subsequent loops at most 0.3.

The pattern persists with facts and demonstrations. Ouro-1.4B answers two-hop fictional questions at 45\%, 64\%, 78\%, and 82\% over one to four loops, and three-hop questions at no more than 56\%. Qwen3-1.7B-Base obtains 65\% and 48\% in one pass. Four solved programs fix Ouro's output format (one-line exact accuracy 3\% to 100\%) while reach stays at most 3.3. Huginn's one-line accuracy is only 55--65\% after one to four recurrences with two chains; reach is 1.4--1.6 after 8, 16, and 32, and two-line accuracy stays at 68--70\% from eight onward. Four demonstrations leave reach at 1.6 after sixteen recurrences.

\newpage
Restoration in Ouro retains 30 of 66 sampled three-line pairs for which both answers are correct and a redirected pointer flips the answer. At least 87\% of each pointer's effect remains at its source through two loops. In layers 9--13 of loop 3, the queried-line pointer's effect falls from 0.90 to 0.30 and the pointer one line earlier from 0.89 to 0.49; loop 4 transfers the remainder. Blocking query attention to pointer assignments in loop 3 or 4 costs 16--24 points, versus at most ten in loop 1 or 2; blocking every loop brings all chains to chance. The reading range spans 3.9--5.7 layers per loop, averaging 4.9.

Ouro's program relay reaches line 3 in loop 1, line 4 only marginally (read-outs 0.80, 0.73, and 0.71 in three samples), and never line 5 in six loops. Huginn reaches line 2 in recurrence 2 and line 3 in recurrence 4, but never line 4 in sixteen (at most 0.65). The chain-selective attention is limited to the parent. Ouro's strongest heads are layer 10 head 13, layer 9 head 12, and layer 10 head 5, with parent attention 0.33--0.37; Huginn's are layer 3 head 34 and layer 2 head 24. Attention two or more lines earlier differs between the own and other chain by at most 0.02. The late query takes additional steps while the program's short relay remains stalled.

\section{What the LoRA unlocks}\label{app:interventions}\label{app:accuracy}

\subsection{Accuracy, format, and model controls}
\begin{figure}[!htb]\centering
\includegraphics[width=\textwidth]{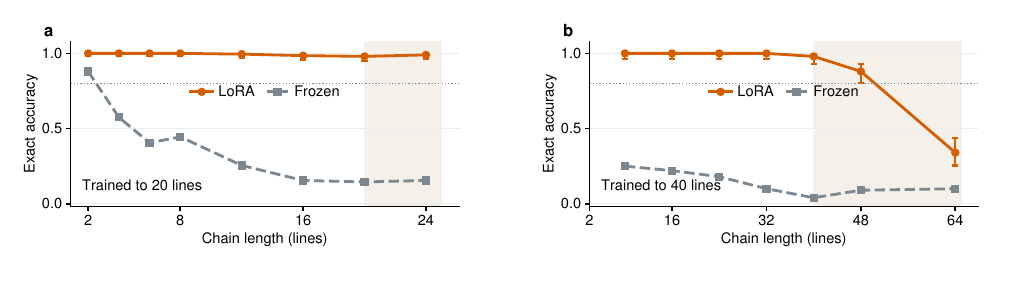}
\caption{\textbf{The same Qwen3-8B weights answer much longer programs after an early LoRA.} Exact next-token accuracy on two-chain, level-order programs. (a) A LoRA trained up to 20 lines, evaluated on 200 programs per length. (b) A separate LoRA trained up to 40 lines with a larger name vocabulary, evaluated on 100. Shading marks extrapolation beyond the training range; bars are 95\% Wilson intervals over programs, not training seeds.}\label{fig:accuracy_std}
\end{figure}

The layer-14 Qwen3-8B LoRA gives exact accuracy 98.5\%, 98.0\%, and 99.0\% at 16, 20, and 24 lines, compared with 15.5\%, 14.5\%, and 15.5\% frozen. Separate training seeds score 94.7--97.5\% at 24 lines. The longer-trained LoRA answers 40- and 48-line programs at 98\% and 88\%, versus 4\% and 9\% frozen. Its reach is 50 lines.

\begin{table}[!htb]\centering\small
\caption{\textbf{The Qwen3-8B LoRA transfers across assignment orders and chain counts.} Exact accuracy (\%), 200 programs per cell. Training uses two chains in level order up to twenty lines; dashes denote unevaluated cells.}\label{tab:accuracy}
\begin{tabular}{llrrrrrr}
\toprule
Chains, order & Model & 4 lines & 8 & 12 & 16 & 20 & 24 \\
\midrule
2, level & frozen & 57.5 & 44.5 & 25.5 & 15.5 & 14.5 & 15.5 \\
2, level & LoRA & 100.0 & 100.0 & 99.5 & 98.5 & 98.0 & 99.0 \\
2, interleaved & frozen & 60.0 & 44.0 & 33.0 & 21.5 & 18.0 & 17.0 \\
2, interleaved & LoRA & 99.5 & 95.0 & 89.5 & 82.0 & 85.5 & 76.5 \\
3, level & frozen & 48.5 & 15.5 & 8.0 & 8.0 & -- & -- \\
3, level & LoRA & 97.0 & 83.5 & 71.5 & 58.0 & -- & -- \\
\bottomrule
\end{tabular}

\end{table}

Changing to randomly interleaved assignments preserves much of the gain (Table~\ref{tab:accuracy}): 89.5\% exact accuracy at twelve lines and 76.5\% at 24. With a third chain, eight-line accuracy is 83.5\%. A separate three-chain LoRA trained on code-style programs up to five lines transfers to JavaScript assignments (\texttt{const B = A;}, five-line exact accuracy 41\% to 99\%) and English assignments (\texttt{B means the same as A.}, 0\% to 94\%). Across models, the best tested LoRA gives at least 24 lines in Llama-3.2-3B, Qwen3-4B, Llama-3.1-8B, OLMo-3-7B, Qwen3-8B, Gemma-3-12B, Qwen3-14B, and OLMo-3-32B; reaches are 22 in Qwen3-1.7B, 13 in Qwen3-0.6B, and 6 in Llama-3.2-1B.

\subsection{Alternative interventions at the same layer}
The comparison in Table~\ref{tab:interventions} trains every intervention on the same two-chain, level-order distribution up to 20 lines: 1,200 steps, batches of 16, answer cross-entropy, and a WikiText penalty of weight one. Evaluation uses 200 fresh programs per length. A rank-8 projection LoRA changes all seven projections of layer 14 (query, key, value, output, gate, up, and down) at learning rate $3\cdot10^{-4}$. It uses about nine times the residual-stream LoRA's parameters and attains 99.0\% and 90.0\% exact accuracy at 16 and 24 lines.

The low-rank FLAS-style flow \citep{jin2026flas} operates on the RMS-normalized state with forward Euler updates
\begin{equation}
 z\leftarrow z+\frac{T}{N}B\,\operatorname{silu}(Az+e(t)).
\end{equation}
The time embedding $e(t)$ is learned; $T\sim U[0.5,2]$ during training and $T=2$ at evaluation. Three integration steps give 99.5\% and 94.0\% at 16 and 24 lines with the LoRA's parameter count; one step gives 95.0\% and 79.0\%. This is a parallel representation-intervention method.

The FLAS flow block follows its released implementation without a concept encoder because there is one task. A time embedding precedes a gated MLP of intermediate width 12,288; the optional causal self-attention phase comes from the first FLAS version and is omitted in later versions. Each phase uses RMSNorm before and after, with a per-channel residual gate initialized to 0.1. The velocity is the block output minus its input, integrated over $N$ steps at learning rate $10^{-4}$. The 168-million-parameter block gives 99.5\% and 90.5\%; adding self-attention (235 million) gives 99.0\% and 98.5\%; omitting the time embedding gives 92.0\% and 70.0\%. Sharing one LoRA across layers 6, 10, and 14 gives 100\% and 94.5\%.

All these early changes extend the chain. At layer 26, the LoRA, low-rank flow, FLAS block, and projection LoRA learn to emit a root value but choose between chains near chance: 52.5--56.0\% at 24 lines, with chance 50\%. Table~\ref{tab:interventions} also reports the one-step block and text-distribution controls. Its residual-stream LoRA row is a separate comparison evaluation from the headline accuracy experiment.

\begin{table}[!htb]\centering\small
\caption{\textbf{Several forms of early intervention extend reference chains.} Exact accuracy (\%) on two-chain programs, 200 per length, with 50\% root-choice chance. Parameter counts denote trained parameters. Text KL is the divergence from frozen WikiText predictions per token, averaged over the final 200 training steps. Layer-26 interventions are after the working region.}\label{tab:interventions}
\resizebox{\textwidth}{!}{\begin{tabular}{llrrrrr}
\toprule
Change & Where & Parameters & Text KL & 8 lines & 16 lines & 24 lines \\
\midrule
none (frozen) & & 0 & 0 & 36.0 & 19.5 & 17.5 \\
LoRA: $h \leftarrow s\,h + BAh$ & rank 8, layer 14 & 66K & 0.005 & 100.0 & 99.0 & 96.5 \\
the same LoRA at three layers & layers 6, 10, 14, shared & 66K & 0.006 & 100.0 & 100.0 & 94.5 \\
low-rank flow (FLAS-style), 3 steps & rank 8, layer 14 & 66K & 0.003 & 100.0 & 99.5 & 94.0 \\
low-rank flow, 1 step & rank 8, layer 14 & 66K & 0.002 & 98.0 & 95.0 & 79.0 \\
FLAS flow block, 3 steps & MLP + time, layer 14 & 168M & 0.012 & 99.5 & 99.5 & 90.5 \\
\quad with self-attention & layer 14 & 235M & 0.010 & 100.0 & 99.0 & 98.5 \\
\quad without time embedding & layer 14 & 151M & 0.006 & 99.0 & 92.0 & 70.0 \\
\quad 1 step & layer 14 & 168M & 0.017 & 99.5 & 98.5 & 95.5 \\
projection LoRA (all projections of one layer) & rank 8, layer 14 & 606K & 0.002 & 99.5 & 99.0 & 90.0 \\
\midrule
\multicolumn{7}{l}{\emph{at layer 26, after the placement limit (LoRAs work up to layer 20):}} \\
LoRA & rank 8, layer 26 & 66K & 0.012 & 49.5 & 53.5 & 54.5 \\
low-rank flow, 3 steps & rank 8, layer 26 & 66K & 0.004 & 49.5 & 48.0 & 54.0 \\
FLAS flow block, 1 step & MLP + time, layer 26 & 168M & 0.049 & 49.0 & 50.5 & 56.0 \\
projection LoRA (all projections of one layer) & rank 8, layer 26 & 606K & 0.002 & 51.5 & 48.5 & 52.5 \\
\bottomrule
\end{tabular}
}
\end{table}

\subsection{Loop count, training length, and transfer}
For Ouro-1.4B, the every-loop layer-6 LoRA reaches 1.9, 7.1, and 17 lines after one, two, and three loops. It solves every tested length up to 24 after four, six, and eight loops; evaluating longer programs with unseen two-letter names gives 27, 37, and 41. The frozen model reaches at most 2.5. Training the longer-trained LoRA on up to 40 lines with the 369-name pool gives 7.4, 25, and 60 lines after two, three, and four loops, about 146 after six, and at least 160 after eight. Eight-loop accuracy on 160-line programs is 87\%, at four times the maximum training length. The observed reach gains per loop grow before saturating: about 5, 10, and 10 lines in loops 2--4 for the standard LoRA, and 17, 35, and then about 43 in loops 3--6 for the longer-trained LoRA. These gains apply within a useful loop range: at twelve loops, accuracy falls at every tested length (91\% at four lines, 68\% at 64), consistent with overthinking in recurrent models \citep{kohli2026loop}. Frozen accuracy is at chance from eight lines onward at every loop count.

Huginn's LoRA trained to twelve lines reaches 2.2, 3.4, 6.6, 9.7, 12.3, and 16.8 after three through eight recurrences, and 25.0 after twelve. With the 527-name pool, text penalty, and training up to 24 lines, reach rises to 31 after eight and 58 after sixteen (95\% interval 37--64), or 2.4 times the maximum training length. Reach subsequently falls to 30 and 13 after 32 and 64 recurrences. Both looped families peak at about twice their training recurrence count. With the larger Huginn vocabulary, frozen accuracy is at chance from two lines onward.

Applying the Ouro LoRA in the first loop alone still gives 9.5 lines after two loops, 21.5 after three, and 24--26 after four through eight. These evaluations reach the transition to unseen two-letter names. The program read-outs locate the computation in the unmodified later loops: loop 1 with the LoRA makes lines 2--5 readable, loop 2 without it makes lines 6--11 readable in layers 7--13, and loop 3 resolves the remaining sixteen-line chain.

Seed and distribution controls retain the improvement. Two additional every-loop seeds show the same growth (Table~\ref{tab:reach}). A second longer-trained LoRA seed reaches 52 after four loops, compared with 60 for the first, and answers 96-line chains at 94\% and 98\% after six and eight. A fixed steering vector at layer 6 has 2,049 parameters and reaches 2.9, 6.1, and 7.1 after two through four loops. Randomly interleaved sixteen-line chains score 99\% at four loops with the LoRA versus 46\% frozen. A LoRA trained on two chains transfers without retraining to three, reaching 5.5, 13.0, and at least 16 after two, three, and four loops, versus 2.6 frozen. A longer-trained LoRA trained on two to four chains reaches at least 64 after four loops on both two and three chains, and 37 on four. In Ouro-2.6B, a layer-12 LoRA reaches 2.0, 8.6, 18.1, and at least 24 after one through four loops, versus 2.6 frozen.

The training loop count changes the speed of the computation. Training with two loops yields reach 4.4 after one and 25 after two, compared with 1.9 and 7.1 for the four-loop LoRA. It rises slightly to 27 at three, then declines to 24 at six and 22 at eight. Training with one loop yields 12.5 lines in a single 24-layer pass; a second loop reduces reach to about two. Attention follows the same ordering: LoRAs trained with one, two, or four loops attend 4, 2, or 1 lines earlier in their first loop. The two-loop LoRA attends 2 and 6 lines earlier during training loops, versus 1 and 2 for the four-loop LoRA, and less far after them. For comparison, Qwen3-1.7B with a LoRA reaches 17--22 lines in one pass of 28 layers, similar to Ouro-1.4B after three passes of 24.

\newpage
Reusing only the middle of a standard model also helps. Qwen3-8B re-enters layers 14--22 once or twice, with the LoRA at every re-entry, and trains on chains up to 40 lines. At 64 lines, exact accuracy is 34\% for the one-pass LoRA, 66\% for one re-entry, and 92\% for two; at 48 it is 88\%, 90\%, and 97\%. Reach increases from 50 to at least 64; the frozen model answers 64-line programs at 10\%.

\section{The computation carried by the middle layers}\label{app:mechanism}

\begin{figure}[!htb]\centering
\includegraphics[width=\textwidth]{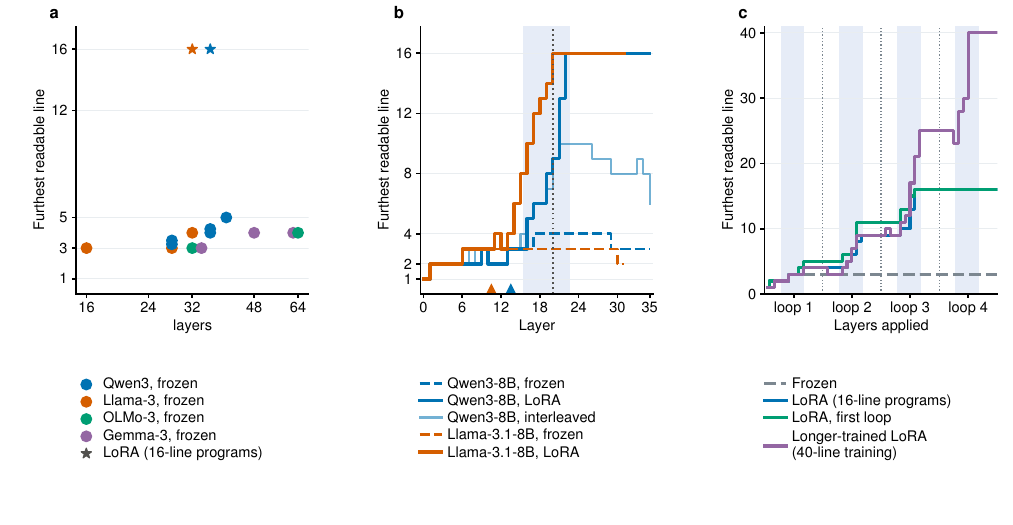}
\caption{\textbf{The relay advances within a short middle range in each pass.} (a) Furthest consecutively readable line versus model depth, frozen on eight-line programs and with LoRA on sixteen-line programs. (b) Layerwise progress in Qwen3-8B and Llama-3.1-8B, with LoRA inputs marked by triangles; shading identifies Qwen3-8B's advancing relay, and the dotted line its placement limit. (c) Progress every second layer of Ouro-1.4B's first four loops: sixteen-line programs for the standard LoRA and forty for the longer-trained LoRA. Shading marks layers 7--15 of each loop.}\label{fig:band}
\end{figure}

\subsection{Where lines become readable}
With the Qwen3-8B LoRA, all sixteen lines become readable by layer 22 in level order, and ten by layer 21 when interleaved. Redirecting line 2's pointer and restoring its state at progressively later lines recovers the original effect; restoration at the query begins to recover it at layer 23 and exceeds half from layer 25. These interventions locate the extended computation inside the program before its result reaches the query.

The relay with LoRA stays at line 3 through layer 15, then reaches lines 5, 6, 6, 8, 9, 13, and 16 at the outputs of layers 16--22. The frozen relay reaches line 4 at 17 and stops. Interleaved chains advance mainly in layers 15--21; Llama-3.1-8B with a layer-11 LoRA advances from line 3 to 16 in layers 13--20. In five standard models with early LoRAs, attention two lines up peaks at 41--50\% of depth.

Ouro-1.4B's relay advances almost entirely in layers 7--15 of each loop. Outside this range it gains at most two lines, apart from the first move to line 2 early in loop 1. On forty-line programs, the longer-trained LoRA adds 3, 5, 16, and 15 lines in loops 1--4 and reaches lines 4, 9, 25, and 40. The final gain is limited by the program length. Ouro-2.6B retains a similarly located and sized range despite its 48-layer body: a layer-12 LoRA advances in layers 7--17 to lines 4, 8, 15, and 24 after one through four loops; the other 37 layers contribute at most two. Huginn advances mostly in the last two layers of its four-layer core. These ranges overlap default query reading: layers 13--22 in Qwen3-8B and layers 9--13 of Ouro's third loop.

\subsection{Attention reaches further as the relay advances}
Using attention difference 0.1 as the threshold, Qwen3-8B attends at most two lines earlier in layers 16--18, then 3, 4, 6, and 7 in layers 19--22. Its relay advances by 2, 1, 0, 2, 1, 4, and 3 lines over those layers, never more than the corresponding attention distance. With interleaved programs, attention reaches two lines earlier in layers 16--18, three to six in 19--22, and ten in 23--24. Heads in layers 20--24 put 62--89\% of their attention on the queried line's own chain and 10--23\% on the other. They distribute it over up to five earlier lines (effective number 1.6--5.0); the same frozen heads devote only 3--21\% to the chain.

In Ouro, the standard LoRA attends 1, 2, 4, and 5 lines earlier across loops 1--4 on 24-line chains; the longer-trained LoRA attends 1, 3, 5, and 6 on forty-line chains. Huginn reaches distances 1, 1, 2, 2, 3, 4, 5, and 5 across its first eight recurrences. Within the middle layers, each two-layer step of Ouro's longer-trained LoRA advances about as far as its attention reaches: about one line per step in loop 1, one or two in loop 2, and four or five in loop 3, where attention reaches five or six. This resembles the growing reach of pointer-doubling constructions \citep{sanford2024logdepth}, but the observed distances increase by one or two rather than doubling, and attention overlaps several earlier lines.

\begin{figure}[!htb]\centering
\includegraphics[width=\textwidth]{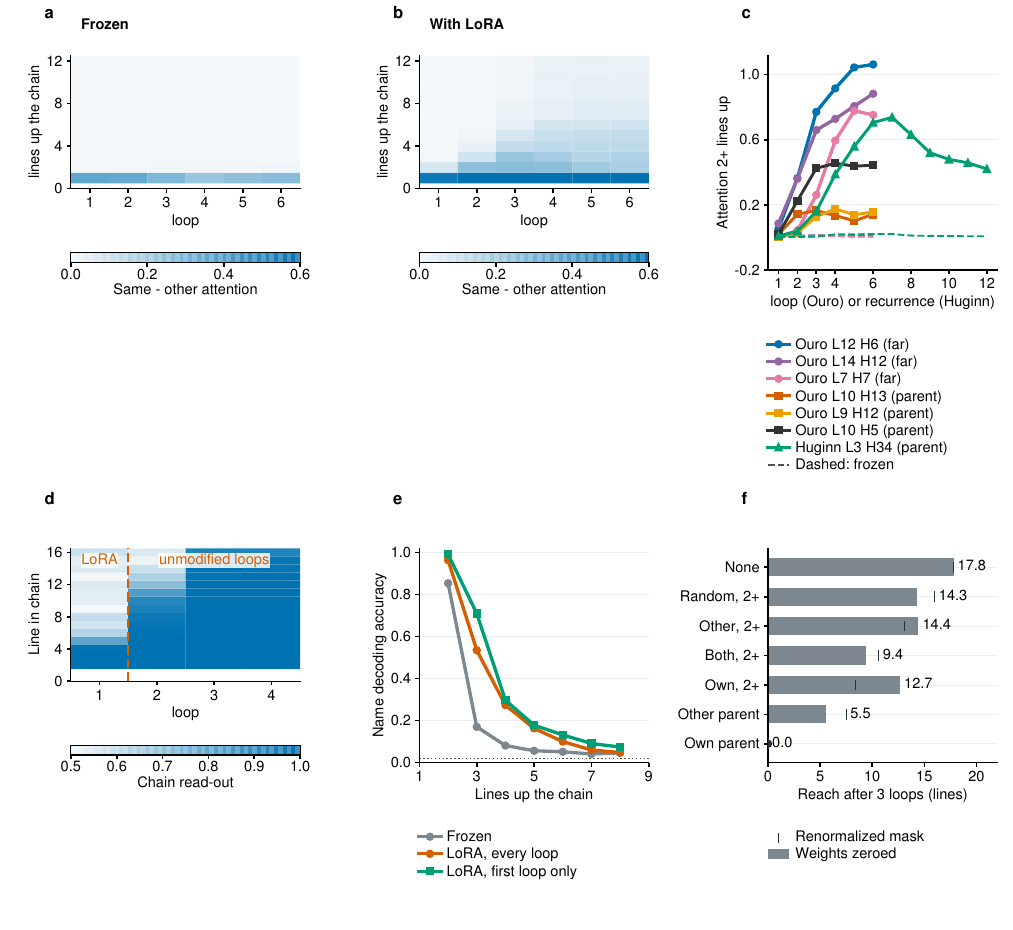}
\caption{\textbf{Frozen heads carry longer reads once the LoRA starts the relay.} Ouro-1.4B unless specified. (a,b) Chain-selective attention by distance and loop on 24-line programs, selecting heads on separate programs. (c) Attention summed over distances of at least two, for the three heads reading furthest with LoRA and the strongest frozen parent-reading heads; solid curves show LoRA and dashed curves the frozen model. (d) Chain read-outs with the LoRA only in loop 1. (e) Decoding ancestor names at lines not yet readable, in loop 2. (f) Three-loop reach after removing attention to specified assignments in every layer and loop: bars zero weights after softmax; ticks renormalize.}\label{fig:mech_loop}
\end{figure}

\subsection{Which heads carry the relay}
The Ouro heads that make longer reads with the LoRA mostly attended no further than the parent when frozen. Layer 12 head 6, layer 14 head 12, and layer 7 head 7 reach total attention differences 0.78--1.06 over distances of at least two in loops 4--6, versus at most 0.02 frozen. Their attention spreads: layer 12 head 6 in loop 4 assigns 0.17, 0.19, 0.20, 0.14, 0.11, and 0.08 at distances one through six. The strongest frozen parent-reading heads themselves read less far with the LoRA. When the LoRA acts only in loop 1, later unmodified loops still read 2, 3, 5, and 4 lines earlier over loops 1--4. In Huginn, the same frozen parent-reading head, layer 3 head 34, takes over longer reads: its attention beyond the parent rises from 0.01 to 0.74 between recurrences 1 and 7.

Head ablations test these associations. In Qwen3-8B with LoRA on interleaved chains, removing the ten parent-reading heads changes choice accuracy at 8, 12, and 16 lines from 95\%, 87\%, and 89\% to 55\%, 46\%, and 53\%; removing ten random heads in the same layers leaves 94\%, 79\%, and 83\%. Removing the eight longer-reading heads gives 79\%, 68\%, and 65\%, compared with 95\%, 87\%, and 90\% for eight random heads.

In Ouro, removing the seven longest-reading heads in every loop leaves reach 8.5 after three loops and 12.0 after four. Twenty random matched-layer draws leave 12.0--17.4 and 22.0--at least 24 (medians 15.8 and 24). Removing the three strongest frozen parent-reading heads (layer 10 heads 13 and 5, layer 9 head 12) leaves 8.5 and 13.7, compared with 9.9--17.5 and 18.7--at least 24 for twenty random draws (medians 15.4 and 24). The same targeted cut lowers frozen two-line accuracy from 91\% to 75\%.

\subsection{What the representations carry}
A probe over variable names shows that each line accumulates names of its ancestors. In Qwen3-8B with LoRA, names defined three to seven assignments earlier become decodable in distance order at layers 17--23, with accuracies from 0.41 down to 0.14 against chance 0.03. Frozen decoding beyond two lines never exceeds 0.19.

In Ouro with 52 names, frozen pointers carry names one and two lines earlier at accuracy 0.89 in loop 1, while names three lines earlier reach at most 0.16 (chance 0.02). The LoRA adds the name three lines earlier at 0.51 in loop 1, and names four, five, and six lines earlier at 0.29, 0.17, and 0.10 in loop 2. These names arrive before chain membership becomes readable. In loop 2, unreadable lines decode names three through six earlier at 0.54, 0.27, 0.16, and 0.10, compared with 0.48, 0.30, 0.18, and 0.11 on readable lines. The unmodified loops after a first-loop-only LoRA give 0.71, 0.30, 0.18, and 0.13. Blocking parent attention throughout loop 2 keeps the four- to six-line names near their loop-1 values: 0.18, 0.09, and 0.05, versus 0.13, 0.06, and 0.04 after loop 1 and 0.29, 0.17, and 0.10 unblocked. Once lines are readable, names fade; three-line-name accuracy is 0.28 by loop 4.

The LoRA initiates this distributed computation. Training and applying it only at program tokens in Qwen3-8B gives 82\% choice accuracy on three interleaved sixteen-line chains, compared with 78\% when applied everywhere; restricting it to query tokens adds only two to four lines. Although the low-rank part of the LoRA writes in the eight-dimensional span of $B$, chain membership is no more readable there than in a random eight-dimensional subspace. Removing that span from later states leaves read-out near 95\%. The relay develops in the wider residual stream. The query then mainly reads its result: blocking Ouro query attention to pointer assignments in any one of loops 1--3 leaves eight- and sixteen-line accuracy at 98--100\%; blocking loop 4 leaves 100\% and 73\%.

\section{Attention cuts and their controls}\label{app:masks}

\subsection{Parent attention and longer reads}
The main Ouro cuts zero attention weights after softmax, leaving all remaining weights unchanged and leaving query attention intact. Each program line loses attention to the specified assignments in every layer and loop. Removing parent attention stops the relay at every tested length and loop count; frozen two-line accuracy also loses 18 points. Removing attention two or more lines earlier in both chains leaves frozen three-line accuracy almost unchanged (67\% without this attention, 65\% with it) but reduces three-loop reach with LoRA from 17.8 to 9.4.

Cuts confined to layers locate the necessary computation. Removing attention in layers 16--23 loses nothing. Removing longer reads in layers 7--15 leaves reach 10.1, versus 15.3 when removed in layers 0--6. Parent attention is needed both before and during the middle range: removing it in layers 0--6 stops the relay, restricting that cut to loop 1 leaves 9.1 lines, and removing it in layers 7--15 leaves 5.0.

The corresponding Qwen3-8B parent cut renormalizes attention. Applied from the LoRA through the relay, layers 14--22, it lowers six-, eight-, and twelve-line accuracy to 53\%, 48\%, and 55\%. Cutting the other chain's corresponding assignments leaves 99\%, 93\%, and 90\%; cutting parent attention only after the relay, in layers 23--29, leaves 100\%, 100\%, and 98\%. In Huginn, a renormalizing cut beyond the parent lowers reach from 16.2 to 10.7 after eight recurrences, while cutting the parent stops the relay.

\subsection{Which chain supplies the useful attention}
A larger Ouro evaluation uses 300 programs per length and 95\% parametric-bootstrap intervals. At three and four loops respectively, removing all reads two or more lines earlier gives reach 9.7 [9.4, 10.0] and 12.1 [11.3, 12.7], compared with 17.1 [16.5, 17.6] and at least 24 unmasked. Cutting only earlier assignments of the same chain gives 12.8 [11.6, 14.0] and 17.3 [15.8, 18.7]; cutting the other chain gives 14.8 [14.3, 15.7] and at least 24. Cutting an equally large random set drawn from both chains gives 14.1 [13.4, 15.0] and 22.5 [21.6, 23.8]. Thus the own chain matters more after four loops. On two chains, the other chain can also identify a binary membership: hiding it makes some long-chain answers systematically wrong, with 34\% accuracy on 24-line chains at three loops.

Three-chain evaluations separate this ambiguity (200 programs per length). Removing reads beyond the parent on the own chain yields 7.6 [6.7, 8.5] and 7.2 [6.5, 8.4] lines after three and four loops. Removing all such reads yields similarly low reach, 7.6 [6.7, 8.3] and 8.4 [7.6, 9.0]. Cutting the other chains leaves 12.4 [11.7, 12.8] and at least 16; cutting an equally large random set leaves 11.7 [10.6, 12.5] and at least 16. Unmasked reach is 12.7 [11.9, 13.3] and at least 16. Here the useful attention is along the line's own chain.

\subsection{Why renormalization changes the comparison}
Assigning blocked keys $-\infty$ before softmax redirects their attention to remaining keys. With this mask, hiding only the own chain hurts more than hiding both chains (three-loop reach 8.4 versus 10.6), because the retained other chain receives the displaced weight. Under renormalizing cuts, faster relays depend more on reads beyond the parent: the two-loop LoRA falls from at least 24 to 12.8 after two loops, and the longer-trained LoRA from 61 to 16.6 after four. Huginn's corresponding cut, evaluated on 60 programs, gives 10.7 instead of 16.2 at eight recurrences and 16.8 instead of at least 24 at sixteen. Parent cuts stop its relay.

\section{Placement: prediction, uncertainty, and loop timing}\label{app:placement}

\subsection{The last useful input layer}
A placement at layer $a$ works when its reach recovers half the best observed gain over median frozen reach $R_0$:
\begin{equation}
 R(a)\geq R_0+\tfrac12\bigl[\max_b R(b)-R_0\bigr].
\end{equation}
The bracket joins the last working placement to the next tested layer. Qwen3-8B falls from 20.5 to 5.2 lines when the LoRA moves from layer 20 to 21; OLMo-3-7B falls from 21.5 to 6.6 between 12 and 15; Llama-3.1-8B from 16.6 to 2.9 between 13 and 15. Second seeds reproduce the contrasts at 21.8 versus 6.5, 20.9 versus 7.7, and 20.0 versus 3.1. Both smaller Qwen models also lose their extension with depth.

For Qwen3-8B, performance varies less across earlier placements: LoRAs at layers 6--19 answer 75--90\% of 24-line programs using two-letter names unseen in training. At layer 20, the last working placement, the relay restarts locally: attention reaches one line earlier at layer 20, two at 21, and three to five by 24. At layer 21, attention stays near the parent (further attention at most 0.11), and the relay does not start. Late layers can therefore continue a computation initiated earlier even when a LoRA at their input cannot initiate it.

\subsection{Prospective test and its resolution}
The placement preregistration was committed September 26, 2026, at 18:12 EDT, before the first saved held-out result at 18:21. Five development models determined three rules: the frozen cutoff layer, 45\% of depth, and the value-copy layer minus 11.5. The preregistration fixed the working threshold above, errors to bracket midpoints, and success within the bracket expanded by one layer on each side in at least three of four held-out models. It also fixed imputing accuracy one at length one when evaluations begin at length two.

The completed held-out grids are layers 1--8, 10, and 12 for Llama-3.2-1B; 12--26 in steps of two for Qwen3-4B; 14, 18, 21, 24, 27, 30, and 34 for Gemma-3-12B; and 14--34 in steps of four for OLMo-3-32B. No held-out sweep has a failing placement followed by a working one. Qwen3-0.6B's layer-3 placement in the development set falls below threshold, so the bracket describes the end of the useful region, not its beginning.

\begin{table}[!htb]\centering\small
\caption{\textbf{The cutoff predicts three of four held-out placement brackets within the preregistered tolerance.} All entries are layers. The first five models informed the rules; the last four were held out. ``Cutoff'' is the frozen measurement, ``Copy'' the value-copy rule, and checks mark predictions within the observed bracket expanded by one layer on either side.}\label{tab:edges}
\begin{tabular}{lrrrrr}
\toprule
Model & Layers & Limit bracket & Cutoff & 45\% & Copy$-11.5$ \\
\midrule
Qwen3-8B & 36 & 20--21 & 20.5$^\checkmark$ & 16.2 & 20.2$^\checkmark$ \\
OLMo-3-7B & 32 & 12--15 & 15.0$^\checkmark$ & 14.4$^\checkmark$ & 12.7$^\checkmark$ \\
Llama-3.1-8B & 32 & 13--15 & 12.5$^\checkmark$ & 14.4$^\checkmark$ & 13.4$^\checkmark$ \\
Qwen3-1.7B & 28 & 12--14 & 18.0 & 12.6$^\checkmark$ & 13.5$^\checkmark$ \\
Qwen3-0.6B & 28 & 12--14 & 10.0 & 12.6$^\checkmark$ & 13.7$^\checkmark$ \\
Llama-3.2-1B & 16 & 7--8 & 8.0$^\checkmark$ & 7.2$^\checkmark$ & 2.8 \\
Qwen3-4B & 36 & 20--22 & 22.0$^\checkmark$ & 16.2 & 20.6$^\checkmark$ \\
Gemma-3-12B & 48 & 24--27 & 23.0$^\checkmark$ & 21.6 & 27.5$^\checkmark$ \\
OLMo-3-32B & 64 & 22--26 & 19.0 & 28.8 & 30.4 \\
\bottomrule
\end{tabular}

\end{table}

\begin{figure}[!htb]\centering
\includegraphics[width=\textwidth]{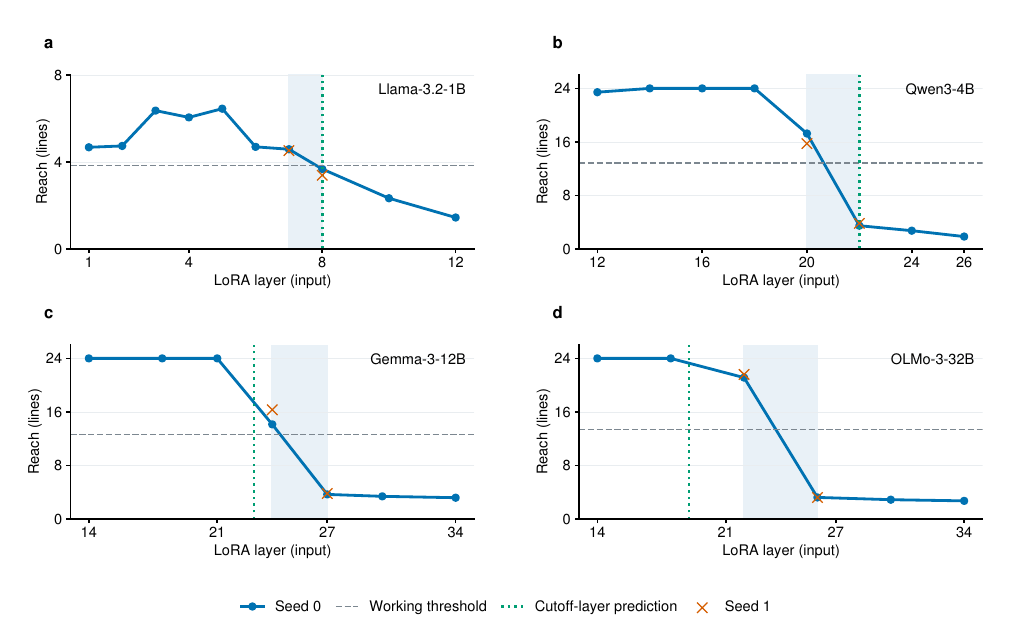}
\caption{\textbf{Held-out sweeps test predictions made from frozen models.} Reach versus LoRA input layer. Circles show the original training seed and crosses a second seed on each side of the bracket. Shading shows the observed bracket, dashed lines the working threshold, and dotted lines the frozen cutoff prediction.}\label{fig:heldout}
\end{figure}

The held-out brackets and cutoff predictions are 7--8 versus 8, 20--22 versus 22, 24--27 versus 23, and 22--26 versus 19, in the order above. Midpoint errors are 0.50, 1.00, 2.50, and 5.00 layers. The first three meet the tolerance, giving mean absolute error 2.25 versus 3.45 for the 45\% rule and 3.37 for the value-copy rule. Across all nine models, cutoff error is 2.22 layers; an after-the-fact constant 48\% of depth gives 2.27. Relative cutoff and relative placement limit correlate at $r=0.69$. The cutoff gives a prior of modest precision; a local sweep establishes the useful range.

Each held-out bracket has a second seed on both sides. Resampling the saved accuracies from their binomials preserves the bracket in at least 99.8\% of 2,000 draws for eight of nine models and in 85.5\% for Llama-3.2-1B. This measures sampling uncertainty conditional on the grid and training seed, not training variation (Table~\ref{tab:flanks}).

\begin{table}[!htb]\centering\small
\caption{\textbf{Independent seeds preserve the observed placement contrasts.} Reach is measured in lines at 80\% exact accuracy. Bracket stability is the fraction of 2,000 binomial resamples retaining the bracket; it conditions on the evaluated grid and seed.}\label{tab:flanks}
\resizebox{\textwidth}{!}{\begin{tabular}{lrrrrrr}
\toprule
Model & Before / after & Seed 0 (before) & Seed 0 (after) & Seed 1 (before) & Seed 1 (after) & Bracket stability \\
\midrule
Qwen3-8B & 20 / 21 & 20.50 & 5.17 & 21.76 & 6.49 & 100.0\% \\
OLMo-3-7B & 12 / 15 & 21.47 & 6.62 & 20.93 & 7.66 & 100.0\% \\
Llama-3.1-8B & 13 / 15 & 16.59 & 2.93 & 20.00 & 3.11 & 100.0\% \\
Qwen3-1.7B & 12 / 14 & 14.86 & 6.74 & -- & -- & 100.0\% \\
Qwen3-0.6B & 12 / 14 & 8.85 & 4.47 & -- & -- & 100.0\% \\
Llama-3.2-1B & 7 / 8 & 4.59 & 3.68 & 4.52 & 3.39 & 85.5\% \\
Qwen3-4B & 20 / 22 & 17.26 & 3.48 & 15.75 & 3.82 & 100.0\% \\
Gemma-3-12B & 24 / 27 & 14.14 & 3.67 & 16.31 & 3.80 & 99.8\% \\
OLMo-3-32B & 22 / 26 & 21.12 & 3.22 & 21.60 & 3.21 & 100.0\% \\
\bottomrule
\end{tabular}
}
\end{table}

\subsection{A later loop can supply the needed middle layers}
In Ouro-1.4B, a layer-10 LoRA inside the middle range reaches 6.2, 17, and at least 24 after two, three, and four loops, similar to layer 6. A layer-20 LoRA, after that range, does nothing in a single loop but can use the next loop: its reach is 4.2, 12.3, and at least 24 after two through four loops, compared with 7.1, 17, and at least 24 at layer 6. Its progress lags by roughly half a loop to a loop.

Applying a LoRA only in the last loop isolates this ordering. Layer 6 still reaches 21.6 after four loops, whereas layer 20 gives 2.6, versus 2.5 frozen, and gains nothing at any tested loop count despite learning the answer format. Ouro-2.6B gives the same distinction with more unused depth: a last-loop LoRA at layer 24 has 24 layers after it but no middle range, adding about two lines from two loops onward and reaching 4.7 after four versus 2.6 frozen. A LoRA at layer 12, within the middle range, adds about seven and reaches 9.6.

\section{Multi-hop questions and fixed-prompt comparisons}\label{app:musique}

\subsection{Fictional facts}
Qwen3-8B's layer-14 LoRA trained on two- to five-hop fictional questions raises exact match from 41.2\% to 97.4\% on 500 matched prompts: a paired gain of 56.2 points [51.8, 60.6]. Across two to five hops, frozen accuracy ranges from 81\% to 26\%, versus 99--100\% with LoRA. Six-hop accuracy, beyond training, rises from 23\% to 88\%. Ouro-1.4B's LoRA trains on one to four hops with three chains (chance 33\%). At four loops it extends reliable answers from two hops (82\% frozen) to six (83\% with LoRA; five-hop accuracy 96\%). At six or eight loops it answers eight-hop questions at 91--92\%, twice the training hops and up to twice the training loop count.

\begin{figure}[!htb]\centering
\includegraphics[width=\textwidth]{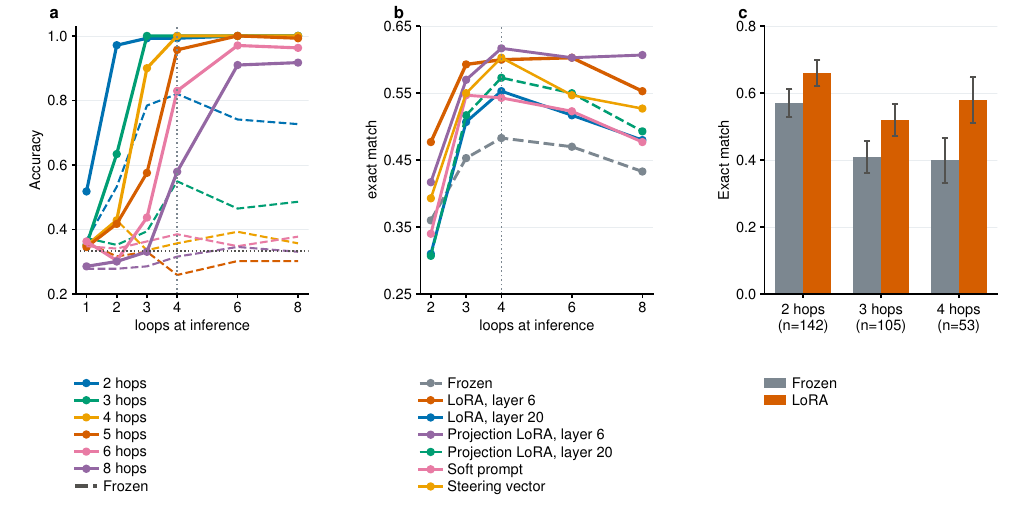}
\caption{\textbf{Early changes improve multi-hop answers in Ouro-1.4B.} (a) Fictional-fact accuracy by hop and loop count, frozen and with a LoRA trained on one to four hops. Three chains give chance 1/3. (b) MuSiQue exact match on 300 development questions using identical prompts for all methods; LoRAs and matched-size projection LoRA are compared at layers 6 and 20, alongside a soft prompt and steering vector. (c) Four-loop exact match by hop count, with binomial standard errors.}\label{fig:downstream_loop}
\end{figure}

\subsection{Prompt matching and uncertainty}
The original paragraph shuffling used a generator seeded by Python's process-specific string hash. Independent processes could therefore see different paragraph orders, shifting frozen exact match by up to four points in Ouro after two loops. Every trained adapter was subsequently evaluated with a fixed hash seed and saved per-question predictions. Comparisons use identical prompts and paired 95\% bootstrap intervals over questions with 2,000 resamples. For multiple training seeds, scores are averaged per question before bootstrapping. These intervals quantify question sampling, not seed variability.

The standard-model setting uses 900 development questions with two to four hops, selection seed one, a fixed demonstration, and supporting paragraphs truncated to 900 characters. Training randomly alternates supporting-only and distractor contexts, uses answer-only teacher forcing, batch eight, and 1,500 AdamW updates with WikiText KL weight one. LoRA learning rate is $10^{-3}$; projection LoRA uses $10^{-4}$ or $3\cdot10^{-4}$. Rank-8 projection LoRA adapts the query, key, value, output, gate, up, and down projections. In Qwen3-8B this trains 21.8 million parameters over all layers, 12.7 million over layers 0--20, or 9.1 million over 21--35; the LoRA trains 65,537.

\subsection{Standard-model MuSiQue controls}
Tables~\ref{tab:qa_std}--\ref{tab:qa_std_other} report the fixed-prompt comparisons. Every tested layer is evaluated on the same 900 development questions; the best early-layer gains summarize this sweep, without a separate held-out split for selecting a layer. Qwen3-8B starts at 52.9\% EM. Early LoRAs at layers 6, 10, and 14 add 11.4, 9.2, and 9.3 points; the layer-14 seeds score 64.3\%, 61.7\%, and 60.4\%. Layer 20 adds 4.8, while 26 and 30 add 0.3 and $-1.6$. The paired layer-6 versus layer-30 difference is 13.0 [10.1, 16.0]. Early projection LoRA gains 11.0 [8.3, 13.8], within 1.8 [$-0.1$, 3.7] of all-layer projection LoRA's 12.8; late projection LoRA gains 0.4 [$-2.3$, 3.3]. At the lower learning rate $10^{-4}$, late projection LoRA still gains only 1.2 [$-1.8$, 4.2].

Equal-width layer ranges rule out parameter count as the explanation. Qwen3-8B projection LoRA on successive nine-layer quarters gains 9.0, 9.4, 4.9, and $-3.1$ points; the first-minus-last difference is 12.1 [8.9, 15.3]. First-half and second-half projection LoRA gain 11.1 and 6.1, a difference of 5.1 [2.7, 7.5]; the second half starts three layers before the program placement limit. OLMo-3-7B starts at 54.4\% and gains 9.4, 6.2, and 5.9 with LoRAs at layers 4, 8, and 12. From layer 16 onward, gains are $-1.1$ to $-3.8$. Early projection LoRA gains 10.6 versus 10.8 for all layers and 0.9 late, with early-minus-late 9.7 [7.2, 12.2]. Llama-3.1-8B starts at 46.8\%, gains 12.9--17.9 up to layer 12, still gains 10.1 at 16 beyond its program limit of 13--15, and at most 2.0 from layer 20. Early projection LoRA gains 20.1, within 0.8 of all-layer projection LoRA's 20.9; late projection LoRA gains 10.7, with early-minus-late 9.4 [6.9, 11.9].

The Qwen3 LoRA at layer 14 recovers about three quarters of all-layer projection LoRA's gain with 65,537 rather than 21.8 million parameters. A 4,097-parameter steering vector there adds 7.8 [5.3, 10.3], only 1.5 [$-0.7$, 3.7] below the LoRA. At layer 6, a LoRA-sized FLAS-style flow adds 9.4 [6.6, 12.3] and the FLAS block 9.3 [6.3, 12.3]. Training a LoRA on single-hop SQuAD in the same way does not improve MuSiQue ($-2.3$ [$-5.2$, 0.7]), supporting a gain beyond answer formatting.

\begin{table}[!htb]\centering\small
\caption{\textbf{MuSiQue comparisons for Qwen3-8B with fixed prompts.} EM and gains average seeds per question; intervals are paired 95\% question-bootstrap intervals on 900 gold-context development questions. Ranges are inclusive. Projection LoRA learning rate is $3\cdot10^{-4}$ unless stated.}\label{tab:qa_std}
\begin{tabular}{llrrl}
\toprule
Model & Intervention & Seeds (EM) & EM & Gain [95\% CI] \\
\midrule
Qwen3-8B & frozen & -- & 52.9 & -- \\
 & LoRA at 6 & -- & 64.3 & +11.4 [+8.6, +14.3] \\
 & LoRA at 10 & -- & 62.1 & +9.2 [+6.1, +12.2] \\
 & LoRA at 20 & -- & 57.7 & +4.8 [+1.9, +7.8] \\
 & LoRA at 26 & -- & 53.2 & +0.3 [-2.2, +3.0] \\
 & LoRA at 30 & -- & 51.3 & -1.6 [-4.2, +1.2] \\
 & LoRA at 14 & 64.3/61.7/60.4 & 62.1 & +9.3 [+6.5, +12.0] \\
 & projection LoRA all & 66.8/64.7 & 65.7 & +12.8 [+9.9, +15.8] \\
 & projection LoRA 0-20 & 63.2/64.6 & 63.9 & +11.0 [+8.3, +13.8] \\
 & projection LoRA 21-35 & 53.1/53.6 & 53.3 & +0.4 [-2.3, +3.3] \\
 & projection LoRA 0-8 & -- & 61.9 & +9.0 [+5.9, +12.2] \\
 & projection LoRA 9-17 & -- & 62.3 & +9.4 [+6.2, +12.9] \\
 & projection LoRA 18-26 & -- & 57.8 & +4.9 [+1.8, +8.0] \\
 & projection LoRA 27-35 & -- & 49.8 & -3.1 [-5.8, -0.4] \\
 & projection LoRA 0-17 & 64.0/64.0 & 64.0 & +11.1 [+8.1, +14.2] \\
 & projection LoRA 18-35 & 60.0/57.9 & 58.9 & +6.1 [+3.4, +9.0] \\
 & projection LoRA all (lr 1e-4) & -- & 61.9 & +9.0 [+5.9, +12.0] \\
 & projection LoRA 21-35 (lr 1e-4) & -- & 54.1 & +1.2 [-1.8, +4.2] \\
 & steering vector at 14 & -- & 60.7 & +7.8 [+5.3, +10.3] \\
 & SQuAD LoRA at 14 & -- & 50.6 & -2.3 [-5.2, +0.7] \\
 & FLAS flow at 6 & -- & 62.3 & +9.4 [+6.6, +12.3] \\
 & FLAS block at 6 & -- & 62.2 & +9.3 [+6.3, +12.3] \\
\bottomrule
\end{tabular}

\end{table}
\begin{table}[!htb]\centering\small
\caption{\textbf{MuSiQue comparisons for OLMo-3-7B and Llama-3.1-8B.} Same 900-question protocol as Table~\ref{tab:qa_std}; projection LoRA learning rate is $10^{-4}$.}\label{tab:qa_std_other}
\begin{tabular}{llrrl}
\toprule
Model & Intervention & Seeds (EM) & EM & Gain [95\% CI] \\
\midrule
OLMo-3-7B & frozen & -- & 54.4 & -- \\
 & LoRA at 4 & -- & 63.9 & +9.4 [+6.2, +12.6] \\
 & LoRA at 8 & -- & 60.7 & +6.2 [+3.1, +9.3] \\
 & LoRA at 12 & -- & 60.3 & +5.9 [+2.6, +9.2] \\
 & LoRA at 16 & -- & 53.3 & -1.1 [-4.2, +2.1] \\
 & LoRA at 20 & -- & 52.4 & -2.0 [-4.8, +0.8] \\
 & LoRA at 24 & -- & 50.7 & -3.8 [-6.8, -0.9] \\
 & LoRA at 28 & -- & 51.1 & -3.3 [-6.3, -0.7] \\
 & projection LoRA all & -- & 65.2 & +10.8 [+7.9, +13.6] \\
 & projection LoRA 0-14 & 66.0/64.1 & 65.1 & +10.6 [+7.7, +13.7] \\
 & projection LoRA 15-31 & 55.2/55.6 & 55.4 & +0.9 [-1.9, +3.6] \\
\midrule
Llama-3.1-8B & frozen & -- & 46.8 & -- \\
 & LoRA at 4 & -- & 64.7 & +17.9 [+14.4, +21.4] \\
 & LoRA at 8 & -- & 59.7 & +12.9 [+9.3, +16.2] \\
 & LoRA at 12 & -- & 63.7 & +16.9 [+13.8, +20.1] \\
 & LoRA at 16 & -- & 56.9 & +10.1 [+7.0, +13.1] \\
 & LoRA at 20 & -- & 48.1 & +1.3 [-1.9, +4.3] \\
 & LoRA at 24 & -- & 47.3 & +0.6 [-2.7, +3.9] \\
 & LoRA at 28 & -- & 48.8 & +2.0 [-0.8, +4.9] \\
 & projection LoRA all & -- & 67.7 & +20.9 [+17.8, +24.0] \\
 & projection LoRA 0-14 & 67.3/66.3 & 66.8 & +20.1 [+17.1, +23.1] \\
 & projection LoRA 15-31 & 56.7/58.2 & 57.4 & +10.7 [+7.8, +13.6] \\
\bottomrule
\end{tabular}

\end{table}

\subsection{Ouro results by loop, hop, and intervention}
The Ouro development set has 300 questions: 142 two-hop, 105 three-hop, and 53 four-hop, with gold paragraphs. With the LoRA applied in every loop, the evaluation uses loop counts 2, 3, 4, 6, and 8. Frozen EM is 36.0\%, 45.3\%, 48.3\%, 47.0\%, and 43.3\%; the layer-6 LoRA gives 47.7\%, 59.3\%, 60.0\%, 60.3\%, and 55.3\%. Paired gains are 11.7 [6.3, 16.7], 14.0 [8.3, 19.7], 11.7 [6.0, 17.3], 13.3 [8.0, 19.0], and 12.0 [6.3, 17.3]. At four loops, two-hop EM rises from 57\% to 66\% and four-hop from 40\% to 58\%. F1 improves less than EM, so part of the gain concerns answer form. An earlier run without fixed hash seeds, using two distractor paragraphs, gives 27.7\% to 38.7\% at two loops and 39.0\% to 54.3\% at four.

A matched-size rank-4 projection LoRA on layer 6's query and value projections has 32,768 parameters and gives 41.7\%, 57.0\%, 61.7\%, 60.3\%, and 60.7\% EM. Residual-stream LoRA minus projection LoRA differences are 6.0 [1.7, 10.7], 2.3 [$-2.0$, 6.7], $-1.7$ [$-5.7$, 2.0], 0.0 [$-4.0$, 4.0], and $-5.3$ [$-9.7$, $-1.0$]. Thus projection LoRA gains 5.7 points at two loops and 11.7--17.3 from three onward.

Applied in every loop at layer 20, the LoRA changes EM by $-5.0$ [$-9.3$, $-0.7$], 5.3, 7.0, 4.7, and 4.7 points over the same loop counts. Layer 6 exceeds it by 16.7, 8.7, 4.7, 8.7, and 7.3, with every interval above zero. Applied in the last loop only, the layer-6 LoRA adds 6.3 points at two, four, and six loops, with intervals above zero; the layer-20 LoRA adds 2.0, $-1.3$, and $-1.3$, with intervals around zero. Projection LoRA at layer 20 gains $-5.3$, 6.3, 9.0, 8.0, and 6.0. Layer-6 projection LoRA exceeds it by 11.0 [6.3, 15.7], 5.3, 4.3, 5.3, and 11.3 [6.0, 16.7].

At four loops, matched application schedules separate the placement comparison: applying the LoRA in every loop gives 60.0\% EM at layer 6 and 55.3\% at layer 20, compared with 48.3\% frozen. The gains are 11.7 [6.0, 17.3] and 7.0 [1.0, 13.0]. Applying it only in the last loop gives 54.7\% at layer 6, a gain of 6.3 [1.0, 11.7], and 47.0\% at layer 20, a change of $-1.3$ [$-6.3$, 3.7]. An every-loop layer-20 LoRA precedes later loops' middle layers; a last-loop layer-20 LoRA does not.

A 16-token soft prompt (32,768 parameters) gains $-2.0$, 9.3, 6.0, 5.3, and 4.3 points, leaving LoRA-minus-prompt differences 13.7, 4.7, 5.7, 8.0, and 7.7. A layer-6 steering vector (2,049 parameters) gains 3.3, 9.7, 12.0, 7.7, and 9.3; LoRA-minus-vector differences are 8.3 [4.0, 12.7], 4.3, $-0.3$, 5.7 [0.3, 10.7], and 2.7.

\section{When the relay appears during learning}\label{app:transfer}

\begin{figure}[!htb]\centering
\includegraphics[width=\textwidth]{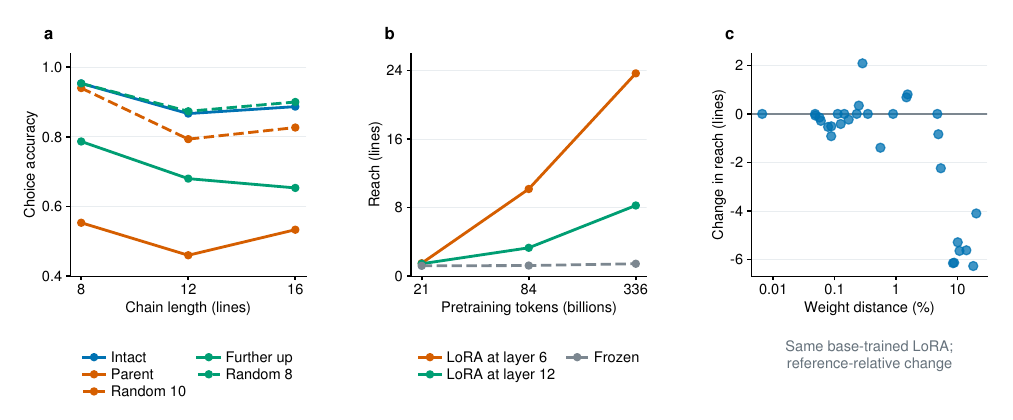}
\caption{\textbf{Targeted head cuts and pretraining checkpoints support a relay in the existing weights.} (a) Qwen3-8B choice accuracy on interleaved programs after removing parent-reading or longer-reading heads, compared with matched-layer random cuts. (b) Reach with identically trained LoRAs at OLMo-3-7B pretraining checkpoints. (c) Change in a base-trained LoRA's reach in post-trained descendants against relative weight distance to the reference checkpoint.}\label{fig:controls_std}
\end{figure}

OLMo-3-7B checkpoints at stage-1 steps 5,000, 20,000, and 80,000 correspond to about 21, 84, and 336 billion tokens, at about 4.2 million tokens per step. With identical LoRA training at layer 6, exact reach is about 1.5, 10.2, and 23.7; frozen reach remains below two. The earliest reach depends on near-chance two-line accuracy and the one-line convention. A dedicated retrieval evaluation shows that the 21-billion-token model with its LoRA already answers one-line lookups at 99\% but two-line chains at 58.5\%. The capacity unlocked between 21 and 84 billion tokens therefore concerns following a link. Layer-12 LoRAs unlock only 3.3 and 8.2 lines at the two later checkpoints, compared with 21.5 in the final model: useful placement also changes during pretraining.

A base-trained LoRA can be transferred unchanged to post-trained descendants. We measure their weight distance on query and MLP down projections at one quarter, one half, and three quarters of depth:
\begin{equation}
 \delta=\left(\frac{\sum_j\lVert W_j-W_j^{(0)}\rVert_F^2}{\sum_j\lVert W_j^{(0)}\rVert_F^2}\right)^{1/2},
\end{equation}
using the nearest candidate reference, either the family's base model or an official post-trained checkpoint. Descendants close in this distance retain the base LoRA's reach; larger changes can reduce it substantially (Figure~\ref{fig:controls_std}c). The fixed base LoRA and answer format remain confounds, so this comparison does not identify a causal effect of the post-training objective.

\section{A relay learned from scratch}\label{app:toy}

\begin{figure}[!htb]\centering
\includegraphics[width=\textwidth]{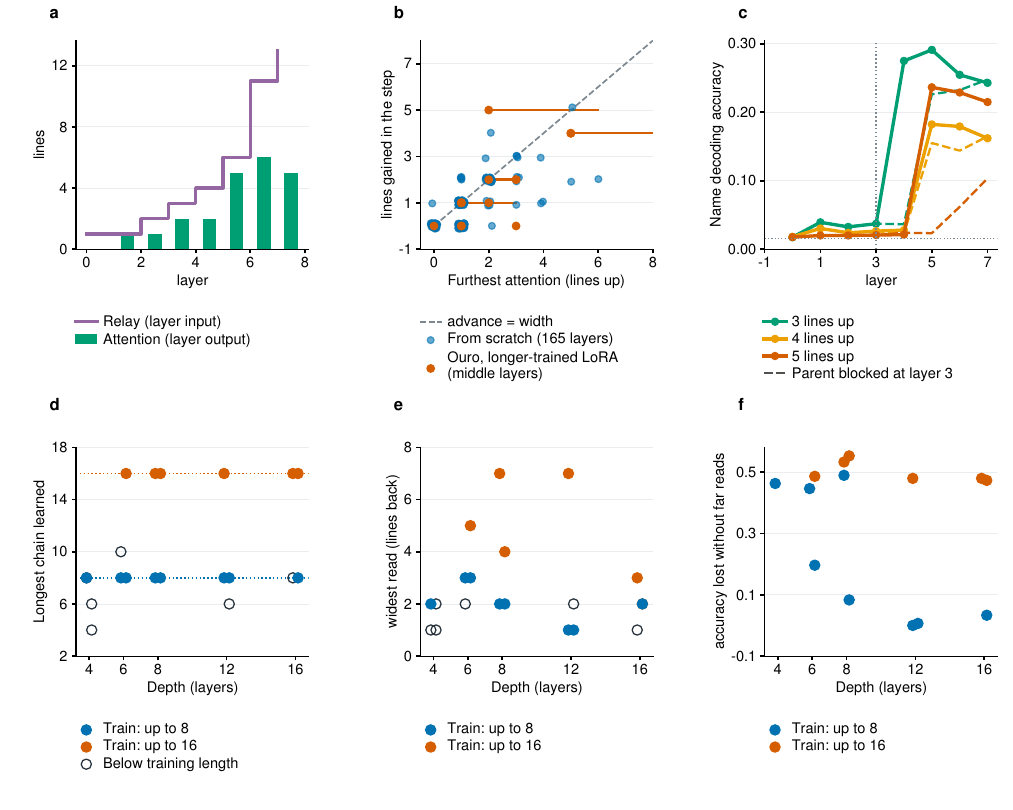}
\caption{\textbf{Small trained transformers learn the same relation between attention and relay progress.} (a) An eight-layer model with a two-layer body applied four times, trained up to sixteen-line chains: readable-line progress at layer inputs and attention distance across each layer. (b) Relay gain versus attention distance across trained models and the two-layer middle steps of Ouro's longer-trained LoRA; bars use attention thresholds 0.1 and 0.05. Integer-valued points overlap: 108 of 165 trained-model steps match, 8 exceed, and 49 fall below the attention distance. (c) Ancestor-name decoding with and without a parent cut at layer 3. (d--f) Models trained up to eight or sixteen lines, two seeds: curriculum length, attention distance, and accuracy lost by cutting reads beyond the parent. Hollow markers indicate models whose evaluated reach falls short of training length; these are omitted from (f).}\label{fig:toy}
\end{figure}

\subsection{Training setup}
The models are pre-norm transformers with rotary positions, width 256, and eight heads. A shared two-layer body is repeated $L/2$ times for $L\in\{4,6,8,12,16\}$. Programs use 64 single-token names, 32 values, two or three chains, level or interleaved order, and one supervised query per chain. AdamW runs 30,000 steps at batch size 256, with learning rate $10^{-3}$, 500 warm-up steps, cosine decay, and weight decay 0.01. The maximum chain length starts at one and grows by one when average training accuracy over the preceding 100 steps exceeds 0.9, up to eight or sixteen. The measurements match those used for pretrained models.

\subsection{Progress and ancestor names}
Across 165 layer steps from 24 models, the relay advances by exactly its attention distance in 108 steps, by more in 8, and by less in 49. The middle two-layer steps of Ouro's longer-trained LoRA follow the same relation. A twelve-layer model that never attends beyond the parent advances exactly one line per layer for seven layers.

The illustrated eight-layer model attends strongly to the parent in layers 1--6 (attention difference 0.89--1.00) and farther in some layers: two lines earlier at layer 4 (0.98), five at layer 5 (0.74), and six at layer 6 (0.55). Ancestor names arrive gradually: two lines earlier at layer 3 with decoding accuracy 0.50 (chance 0.016), three at layer 4 with 0.27, four and five at layer 5, and six at layer 6. Each layer adds one or two earlier names rather than doubling the set.

Removing parent attention only at layer 3 lowers three-line-name decoding at layer 4 from 0.27 to 0.04; it arrives a layer later at 0.23. Five- and six-line names remain at 0.02--0.10 through the remaining layers, versus 0.16--0.24 unblocked. A layer-2 cut delays the two- and three-line names by one to two layers. Conversely, removing attention beyond the parent at layers 5 and 6 leaves name decoding unchanged (five-line name 0.24, six-line name 0.15), while removing it throughout the network reduces sixteen-line accuracy to 0.53. In this model, parent attention brings names to a line, and longer reads use those names to find its chain.

\subsection{Depth changes how far attention must reach}
Among models that learn eight-line training chains, four- to eight-layer models attend two or three lines earlier, twelve-layer models attend only one, and a sixteen-layer model attends two. Cutting attention beyond the parent costs 0.03--0.49 accuracy on six-line chains in models using longer reads and nothing in the twelve-layer models. Six of ten models learn the sixteen-line curriculum: both seeds at eight and sixteen layers, one at six, and one at twelve. At the same depth they attend farther than on the shorter curriculum: five lines at six layers, four to seven at eight and twelve, and two or three at sixteen. Every model that learns this curriculum loses about half its twelve-line accuracy without longer reads (0.47--0.55). Names three or more lines earlier are decodable only in models making such reads.

Three-chain models show both ways of identifying membership. The illustrated eight-layer model and a twelve-layer model trained to sixteen lines relay along their own chain: hiding its earlier assignments lowers reach to about eight and twelve, while hiding other chains changes little. A six-layer model instead uses the other chains' assignments; hiding them makes answers systematically wrong, with 0.13--0.18 accuracy on six- and eight-line chains against chance 0.33.

\section{Complete looped reach estimates}\label{app:reach}

Tables~\ref{tab:reach}--\ref{tab:reach_huginn} report reach estimates and 95\% parametric-bootstrap intervals. Each accuracy cell is resampled from a binomial with its evaluation sample size in 2,000 draws: ordinarily 150 programs, 100 for longer-trained LoRAs or chains longer than 24, and 60 for Huginn and the longer-trained LoRA's 128- and 160-line cells. A lower-bound sign means every tested length was answered; $<$ means the shortest tested length was not. Bootstrap reach estimates are capped at the longest tested length. Thus a censored entry such as $\geq24.0\,[24.0,24.0]$ means both interval endpoints reach the evaluation ceiling, not that the uncensored reach is known without uncertainty. For Ouro, one-line chains count as answered when the shortest tested length is one or two, and LoRAs enter layer 6 unless stated. Its standard LoRAs train on at most twenty lines with single-letter names; evaluations above 26 lines introduce unseen two-letter names. Longer-trained LoRAs train to forty lines with both kinds of name. The mixed-chain longer-trained LoRA trains on two to four chains and is evaluated separately at chance levels one half, one third, and one quarter.

\begin{table}[!htb]\centering\footnotesize
\caption{\textbf{Ouro reach over the first four loops.} Lines at 80\% choice accuracy, with 95\% parametric-bootstrap intervals; the same trained LoRAs continue in Table~\ref{tab:reach_later}.}\label{tab:reach}
\begin{tabular}{lcccc}
\toprule
Ouro-1.4B & $T=1$ & $T=2$ & $T=3$ & $T=4$ \\
\midrule
Ouro-1.4B frozen & 0.0 [0.0, 0.0] & 1.9 [1.7, 2.1] & 2.3 [1.9, 2.6] & 2.5 [2.3, 2.8] \\
steering vector & -- & 2.9 [2.7, 3.2] & 6.1 [5.6, 6.5] & 7.1 [6.8, 7.7] \\
LoRA, every loop (seed 0) & 1.9 [1.7, 2.2] & 7.1 [6.7, 7.7] & 17.0 [16.2, 17.5] & 26.7 [26.1, 27.6] \\
LoRA, every loop (seed 1) & 1.8 [1.6, 2.1] & 7.3 [7.0, 7.8] & 20.0 [18.5, 21.1] & $\geq$24.0 [24.0, 24.0] \\
LoRA, every loop (seed 2) & 2.3 [2.1, 2.5] & 7.0 [6.8, 7.4] & 17.9 [17.2, 18.8] & $\geq$24.0 [24.0, 24.0] \\
LoRA, first loop only & -- & 9.5 [8.9, 10.1] & 21.5 [20.0, 23.2] & 26.3 [25.3, 27.4] \\
LoRA at layer 10 & 1.9 [1.7, 2.1] & 6.2 [5.6, 6.6] & 17.2 [16.1, 18.3] & $\geq$24.0 [24.0, 24.0] \\
LoRA trained with one loop & 12.5 [11.5, 13.2] & 2.2 [1.7, 4.6] & 0.0 [0.0, 4.6] & 3.6 [0.0, 4.3] \\
LoRA trained with two loops & 4.4 [4.1, 4.7] & 25.1 [23.6, 26.1] & 26.9 [26.3, 27.7] & 25.5 [24.6, 26.2] \\
longer-trained LoRA (up to 40 lines) & -- & 7.4 [6.4, 9.3] & 24.6 [22.4, 26.6] & 60.0 [56.0, 66.5] \\
longer-trained LoRA, second seed & -- & $<$8 & 19.9 [18.5, 21.8] & 52.1 [48.0, 55.3] \\
longer-trained LoRA, 2-4 chains: two chains & -- & 10.5 [9.4, 11.5] & 31.4 [28.4, 34.7] & $\geq$64.0 [64.0, 64.0] \\
\quad three chains & -- & 8.4 [8.0, 9.3] & 28.4 [27.2, 30.3] & $\geq$64.0 [58.0, 64.0] \\
\quad four chains & -- & 8.3 [8.0, 9.2] & 24.2 [21.3, 25.3] & 36.6 [30.4, 44.8] \\
\bottomrule
\end{tabular}

\end{table}
\begin{table}[!htb]\centering\footnotesize
\caption{\textbf{Ouro reach with additional inference loops.} Same estimates and conventions as Table~\ref{tab:reach}.}\label{tab:reach_later}
\begin{tabular}{lccc}
\toprule
Ouro-1.4B & $T=6$ & $T=8$ & $T=12$ \\
\midrule
Ouro-1.4B frozen & 2.3 [2.1, 2.5] & 2.2 [1.9, 2.4] & -- \\
steering vector & 6.8 [5.9, 7.6] & 6.3 [5.6, 6.9] & -- \\
LoRA, every loop (seed 0) & 37.4 [36.1, 39.2] & 40.9 [39.1, 43.9] & -- \\
LoRA, every loop (seed 1) & $\geq$24.0 [24.0, 24.0] & $\geq$24.0 [24.0, 24.0] & -- \\
LoRA, every loop (seed 2) & $\geq$24.0 [23.0, 24.0] & $\geq$24.0 [24.0, 24.0] & -- \\
LoRA, first loop only & 24.5 [23.0, 25.7] & 24.0 [19.5, 25.7] & -- \\
LoRA at layer 10 & $\geq$24.0 [24.0, 24.0] & 21.5 [15.1, 24.0] & -- \\
LoRA trained with one loop & -- & -- & -- \\
LoRA trained with two loops & 23.8 [22.4, 25.5] & 21.6 [15.7, 23.4] & -- \\
longer-trained LoRA (up to 40 lines) & 145.5 [132.9, 160.0] & $\geq$160.0 [156.0, 160.0] & 20.0 [7.0, 24.0] \\
longer-trained LoRA, second seed & $\geq$96.0 [96.0, 96.0] & $\geq$96.0 [96.0, 96.0] & -- \\
longer-trained LoRA, 2-4 chains: two chains & $\geq$64.0 [64.0, 64.0] & $\geq$64.0 [64.0, 64.0] & -- \\
\quad three chains & $\geq$64.0 [46.5, 64.0] & $\geq$64.0 [46.0, 64.0] & -- \\
\quad four chains & 41.3 [32.0, 52.4] & 41.3 [32.0, 51.2] & -- \\
\bottomrule
\end{tabular}

\end{table}
\begin{table}[!htb]\centering\footnotesize
\caption{\textbf{Huginn reach by recurrence count.} The standard LoRA trains up to twelve lines and the longer-trained LoRA up to 24; intervals and censoring follow Table~\ref{tab:reach}.}\label{tab:reach_huginn}
\begin{tabular}{lccc}
\toprule
Huginn-0125 & $r=2$ & $r=4$ & $r=8$ \\
\midrule
Huginn-0125 frozen & 0.0 [0.0, 0.0] & 0.0 [0.0, 0.0] & 1.6 [1.4, 2.0] \\
LoRA (up to 12 lines) & 0.0 [0.0, 0.0] & 3.4 [2.7, 4.3] & 16.8 [15.6, 17.6] \\
longer-trained LoRA (up to 24 lines) & $<$2 & 4.6 [3.6, 5.4] & 30.5 [27.8, 34.8] \\
\midrule
Huginn-0125 & $r=16$ & $r=32$ & $r=64$ \\
\midrule
Huginn-0125 frozen & 1.4 [0.0, 2.0] & 1.5 [1.1, 2.0] & -- \\
LoRA (up to 12 lines) & $\geq$24.0 [24.0, 24.0] & $\geq$24.0 [23.1, 24.0] & -- \\
longer-trained LoRA (up to 24 lines) & 57.6 [37.3, 64.0] & 30.0 [15.1, 41.6] & 12.6 [10.3, 16.0] \\
\bottomrule
\end{tabular}

\end{table}

\clearpage
\section{Further connections to prior work}\label{app:related}

\paragraph{Reference chains and entity tracking.} Variable assignments connect this task to entity tracking \citep{kim2023entity} and the variable-tracking task in RULER \citep{hsieh2024ruler}. Tracking entities across state changes \citep{tang2026entitystate} and retrieval-conditioned rebinding \citep{oh2026rebinding} address related questions about how representations preserve and update references. Mechanistic studies of symbolic multi-step reasoning \citep{brinkmann2024symbolic} and variable binding \citep{wu2025binding} provide close precedents. Our distinctive question is how a task-trained edit changes reference following in otherwise frozen pretrained weights. We measure the resulting program relay, follow it through repeated loops, and test where an intervention can initiate it. The from-scratch models test whether the same relation between attention and relay progress holds in a controlled setting.

\paragraph{Recurrent algorithms.} Looped transformers can emulate deeper networks and parallel algorithms such as pointer doubling \citep{sanford2024logdepth,saunshi2025latent}; shallow transformers also learn logarithmic-depth shortcuts for automata \citep{liu2023shortcuts}. Recurrent networks can solve harder instances with more iterations until overthinking sets in \citep{schwarzschild2021learn,bansal2022overthinking}. Studies of pretrained looped models find repeated processing stages \citep{blayney2026stages}, a workspace rebuilt each loop \citep{wang2026jlensloop}, and a preference for serial algorithms under weight tying \citep{zhang2026convergence}. These are distinct from demonstrations of hop extrapolation on parametric tasks \citep{kohli2026loop} or of one hop per loop installed through per-loop supervision \citep{shapiro2026}. Growing and looping offer different forms of iterative computation \citep{kapl2026growing}. Here, the critical distinction is between the short default computation on in-context chains and the longer relay accessible through a small intervention.

\paragraph{Efficient adaptation and added computation.} Small intrinsic dimensions \citep{aghajanyan2021intrinsic} and circuit reuse during fine-tuning \citep{prakash2024entity} motivate studying how little must change. Localized representation fine-tuning also studies where to adapt a model \citep{yin2024lofit}. Alongside LoRA and FLAS interventions, steering vectors add a fixed direction \citep{turner2023steering,rimsky2024caa,zou2023repe}, while prompt tuning learns input representations \citep{lester2021prompt}. Filler tokens and continuous thoughts provide additional computation without written reasoning \citep{pfau2024dots,hao2024coconut}. Single-layer reinforcement learning at 39--47\% of depth can recover full-training gains \citep{zhang2026onelayer}. Our comparisons use intervention form as a control: the focus is where the remaining frozen layers can carry the chain.

\section{Reproducibility}\label{app:repro}
The experiments use synthetic prompts, public models, and public datasets. The project page (\url{https://lunamos.github.io/stop-thinking-too-early/}) links to the code: the experiment scripts, the analysis scripts that compute every table and figure from the experiments' outputs, and a guide from each claim to the script and command behind it.

\end{document}